\documentclass[nohyperref]{article}

\usepackage{microtype}
\usepackage{graphicx}
\usepackage{subfigure}
\usepackage{booktabs} 

\usepackage{hyperref}

\usepackage[accepted]{icml2022}

\usepackage{amsmath}
\DeclareMathOperator*{\argmax}{arg\,max}

\usepackage{amssymb}
\usepackage{mathtools}
\usepackage{amsthm}
\usepackage{multirow}

\usepackage[capitalize,noabbrev]{cleveref}

\theoremstyle{plain}

\theoremstyle{definition}

\theoremstyle{remark}

\usepackage[textsize=tiny]{todonotes}

\icmltitlerunning{Confidence Score for Source-Free Unsupervised Domain Adaptation}

\begin{document}

\twocolumn[
\icmltitle{Confidence Score for Source-Free Unsupervised Domain Adaptation}



\icmlsetsymbol{equal}{*}


\begin{icmlauthorlist}
\icmlauthor{Jonghyun Lee}{snu}
\icmlauthor{Dahuin Jung}{snu}
\icmlauthor{Junho Yim}{comp}
\icmlauthor{Sungroh Yoon}{snu,prof}
\end{icmlauthorlist}

\icmlaffiliation{snu}{Data Science and AI Lab., Seoul National University}
\icmlaffiliation{prof}{
Department of ECE and Interdisciplinary Program in AI, Seoul National University}
\icmlaffiliation{comp}{AIRS Company, Hyundai Motor Group, Seoul, Korea}

\icmlcorrespondingauthor{Sungroh Yoon}{sryoon@snu.ac.kr}

\icmlkeywords{Machine Learning, ICML}

\vskip 0.3in
]



\printAffiliationsAndNotice{}  

\begin{abstract}
\label{0_Abstract}

Source-free unsupervised domain adaptation (SFUDA) aims to obtain high performance in the unlabeled target domain using the pre-trained source model, not the source data.
Existing SFUDA methods assign the same importance to all target samples, which is vulnerable to incorrect pseudo-labels.
To differentiate between sample importance, in this study, we propose a novel sample-wise confidence score, the Joint Model-Data Structure (JMDS) score for SFUDA.
Unlike existing confidence scores that use only one of the source or target domain knowledge, the JMDS score uses both knowledge.
We then propose a Confidence score Weighting Adaptation using the JMDS (CoWA-JMDS) framework for SFUDA.
CoWA-JMDS consists of the JMDS scores as sample weights and weight Mixup that is our proposed variant of Mixup.
Weight Mixup promotes the model make more use of the target domain knowledge.
The experimental results show that the JMDS score outperforms the existing confidence scores.
Moreover, CoWA-JMDS achieves state-of-the-art performance on various SFUDA scenarios: closed, open, and partial-set scenarios.

\end{abstract}

\section{Introduction}
\label{1_Introduction}
Recently, Deep Neural Networks (DNNs)~\citep{lecun2015deep} have successfully demonstrated high performance in various applications. 
However, if the distribution of the training and test data differs, significant performance degradation occurs, which is known as a domain shift~\citep{transferlearningsurvey}.
Unsupervised domain adaptation (UDA) mitigates the domain shift problem using both fully annotated source and unlabeled target data with the assumption that the data distributions in the two domains are slightly different. The UDA task aims to obtain a high target performance using the two domains without the target label.

All conventional UDA methods assume the availability of both the source data and corresponding labels. However, this may be impractical in some cases. First, growing concerns regarding data privacy and security force companies to only release the data. Second, many resources such as GPUs and time are required to train a model when source data are much greater than target data. To address these concerns, source-free UDA (SFUDA) has recently been studied~\citep{shot, 3cgan, bait, sfit}. Instead of accessing the source data, SFUDA assumes that we can access only a pre-trained model using labeled source domain data. 

\begin{figure}[t]
    \centering
    \includegraphics[width=1.0\linewidth]{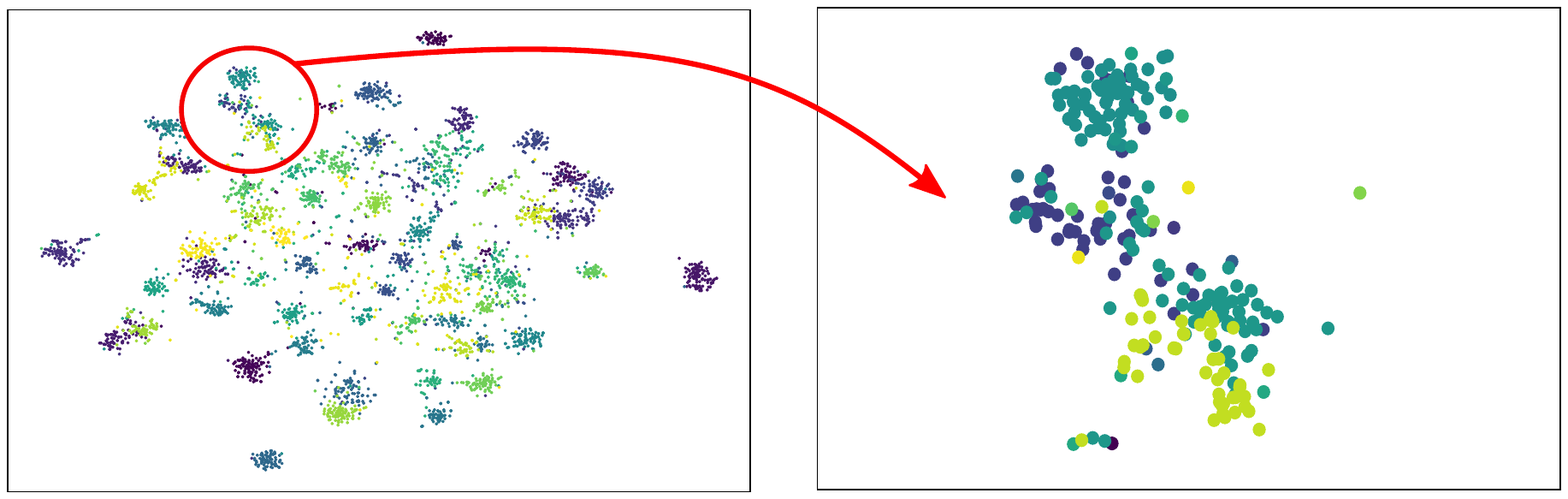}

    \vspace{-0.25cm}
    \caption{(Left) t-SNE plot of the target feature in Ar $\rightarrow$ Rw task on the Office-Home dataset. (Right) Zoomed-in t-SNE plot. The color of samples indicates the ground truth labels.}
    \label{fig:tsne_zoom}
    \vspace{-0.5cm}
\end{figure}

Existing SFUDA methods strictly follow the cluster assumption~\cite{clusterassumption} and use pseudo-labels~\cite{pseudolabel} of target data based on the target feature cluster. 
Because the only data accessible in SFUDA are the target data, the model is updated while preserving its intrinsic cluster architecture.
In other words, existing SFUDA methods train the model so that its decision boundary does not penetrate the target feature cluster.
Figure~\ref{fig:tsne_zoom} shows t-SNE~\cite{tsne} plots to visualize the inherent architecture of the target features obtained by the pre-trained source model.
As shown in Figure~\ref{fig:tsne_zoom}, the samples form their own cluster.
However, according to the right side of Figure~\ref{fig:tsne_zoom}, ground truth labels of samples included in the same cluster are diverse for some clusters.
The model assigns the same pseudo-label to samples in the same cluster based on the cluster assumption, which leads incorrect pseudo-labels.
To robustly learn with incorrect pseudo-labels, samples with low-confidence in their pseudo-label should be suppressed when the model trains~\cite{revisitingconfidencescore}.
The existing methods are limited in that they are vulnerable to incorrect pseudo-labels and confirmation bias~\cite{confirmationbias}, also known as noise accumulation, because it allocates the same weight to all samples regardless of their confidence in the pseudo-labels.

In this study, we propose a novel confidence score for SFUDA, the Joint Model-Data Structure (JMDS) score, to differentiate between sample importance based on the confidence for pseudo-labels.
SFUDA has two components: the pre-trained source model and target data.
The model has knowledge of the source domain, such as class similarity, because it was trained in the source domain.
On the other hand, the target feature distribution obtained from the target data has knowledge of the target domain such as data similarity.
Therefore, the JMDS score should include  knowledge of the model and target feature distribution to fully utilize knowledge of both domains.

The JMDS score consists of two confidence scores, a Log Probability Gap (LPG) score and a Model Probability of Pseudo-Label (MPPL) score, to include both knowledge. 
We first use Gaussian Mixture Modeling (GMM) in the target feature space to obtain the log-likelihood and pseudo-label of each sample.  
The LPG score, the data-structure-wise confidence score, is the gap between the primary and secondary classes of log probability based on GMM.
The MPPL score, the model-wise confidence score, is the probability of the model for a corresponding pseudo-label obtained from GMM.
The LPG and MPPL scores include knowledge of the target and source domains, respectively.
Therefore, to the best of our knowledge, the JMDS score is the first confidence score that includes both source and target domain knowledge.

The objective of the JMDS score is to measure the sample-wise confidence for pseudo-labels at the given model, not the adaptation. 
To use the JMDS score in the learning process for SFUDA, we propose a novel framework, Confidence Score Weighting Adaptation using the JMDS (CoWA-JMDS) framework.
CoWA-JMDS uses the JMDS score as a sample-wise weight and pseudo-labels obtained from GMM.
Also, it uses weight Mixup, which is our proposed variant of Mixup~\cite{mixup}.
Because sample weighting with the JMDS score suppresses low-confidence samples, knowledge of the target feature distribution may not be sufficiently exploited.
Weight Mixup promotes the model make more use of target domain knowledge by mixing low-confidence samples with other samples and considering confidence of the mixed samples.
CoWA-JMDS can be easily extended to open-set and partial-set scenarios with minor modifications.

We evaluated the JMDS score and CoWA-JMDS on various public UDA benchmarks.
First, the JMDS score achieved the best performance in terms of measuring confidence compared to existing confidence scores.
Because performance using the JMDS is better than using the MPPL or LPG alone, it has been experimentally proven that knowledge of both domains is important in SFUDA.
Second, despite its simplicity, CoWA-JMDS outperformed the state-of-the-art SFUDA method on three benchmarks in closed-set scenarios without any auxiliary networks.
CoWA-JMDS also achieved the best performance in open-set and partial-set scenarios.
Through further analysis in Section~\ref{6_Discussion}, we provide an explanation on why the JMDS score is reliable in SFUDA.
Additionally, our proposed weight Mixup improves 3.4\% in terms of the average accuracy on Office-31 dataset than Mixup. It demonstrates that weight Mixup is an effective technique in learning with sample weighting based on confidence scores.
The code is available at \href{https://github.com/Jhyun17/CoWA-JMDS}{https://github.com/Jhyun17/CoWA-JMDS}.
\textcolor{black}{
Our contributions can be summarized as follows:
\begin{itemize}
    \item We propose a novel confidence score, the JMDS score, which considers knowledge of both the source and target domains in SFUDA. 
    \item We propose an SFUDA framework, CoWA-JMDS, which uses JMDS scores as sample-wise weights and weight Mixup which is proposed to exploit more target domain knowledge.
    \item We demonstrate that the proposed JMDS score and CoWA-JMDS achieve state-of-the-art performance on UDA benchmarks. 
\end{itemize}
}

\section{Related work}
\label{2_Related work}
\begin{figure*}[!htb]
    \centering
    \includegraphics[width=0.95\linewidth]{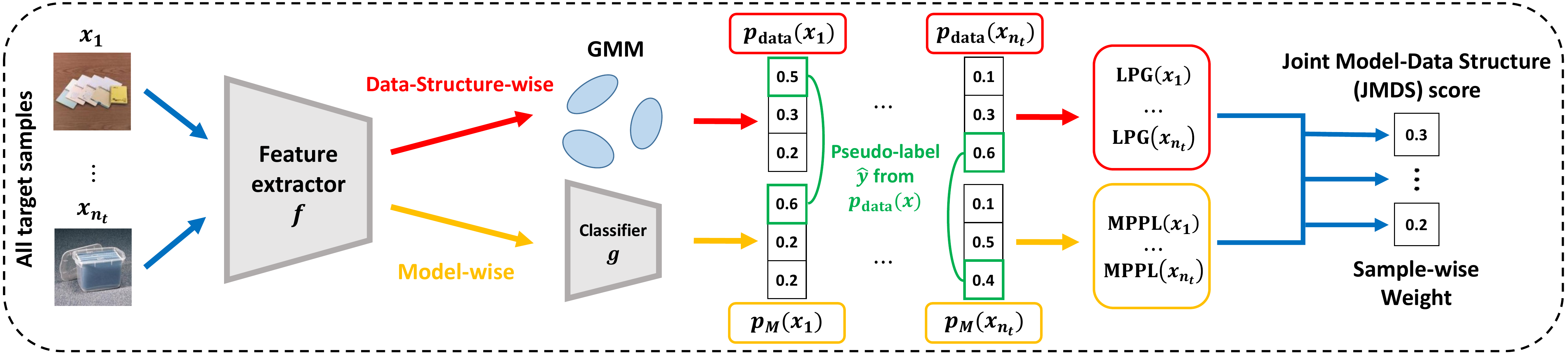}
    \vspace{-0.3cm}
    \caption{Overview of the JMDS score. The JMDS score consists of two scores: The LPG and MPPL scores. The LPG score uses the data-structure-wise probability from GMM, while the MPPL score uses the model-wise probability. The red color indicates the data-structure-wise knowledge and the yellow color indicates the model-wise knowledge. The pseudo-labels of samples are decided by the data-structure-wise probability $p_\textmd{data}(X_t)$.
    }
    \label{fig:JMDS_overview}
    \vspace{-0.3cm}
\end{figure*}

\subsection{Source-free unsupervised domain adaptation}
SFUDA is a more difficult UDA setting than the conventional UDA, where only a source-trained model can be used. Thus, the existing methods for UDA that use source data directly cannot be applied to SFUDA. 

Collaborative Class Conditional Generative Adversarial Networks (3C-GAN)~\cite{3cgan} are based on time-consuming target-style image generation through a conditional GAN. Source-Free Image Translation(SFIT)~\cite{sfit} uses knowledge distillation to translate target images into source style without using source images. Both methods require an auxiliary network. 

Source HypOthesis Transfer (SHOT)~\cite{shot} matches the target feature to a fixed pre-trained source classifier that fine-tunes the feature extractor. SHOT uses self-supervised pseudo-labeling (SSPL) and information maximization loss to balance the pseudo-label and avoid trivial solutions. However, SHOT cannot fully extract the knowledge of the target feature structure because it uses SSPL which does not consider the covariance of each dimension on the feature space and cluster density.

Neighborhood Reciprocity Clustering (NRC)~\cite{nrc} exploits the intrinsic neighborhood structure of the target data in the feature space. It uses the nearest neighbors and an affinity matrix on the target feature space to exploit the knowledge of the target data distribution. However, it is difficult to use existing techniques such as Mixup~\cite{mixup} and data augmentation.

In this study, we propose an SFUDA framework to overcome these weaknesses.
It uses GMM on the feature space to extract knowledge of the target data structure. GMM has the advantage of obtaining a sample log-likelihood compared to other clustering methods.
Additionally, the framework does not require auxiliary networks and can be effectively combined with a variant of Mixup.

\subsection{Confidence score}
\citet{defineconfidencescore} divided confidence scores into two main tasks; ordinal ranking and probability calibration. In this study, we focused on ordinal rankings.
Ordinal ranking is commonly used for selective classification~\cite{selective2, selective1, selective3, selective4}, which is a task that discriminates samples according to their confidence level for labels to avoid low-confidence samples during training. 
In this work, we present for the first time an SFUDA method that uses a confidence score to robustly learn low-confidence samples that are more likely to have incorrect pseudo-labels.

\section{Joint Model-Data Structure (JMDS) score}
\label{3_JMDS}
In SFUDA, we can only use the target data $X_t=\{x_{i}^{t}\}_{i=1}^{n_t}$ and model $M$, not the source data.
The ground truth labels of the target data, $Y_t = \{y_i^{t}\}_{i=1}^{n_t}$, are inaccessible during the learning stage. 
Instead, the pseudo-labels of the target data, $\hat{Y}_t = \{\hat{y}_i^{t}\}_{i=1}^{n_t}$, are utilized during the learning stage. 
The pseudo-labels are obtained through the prediction of the model or the feature distribution.
The model $M$ is pre-trained using labeled source data $X_s=\{x_{i}^{s}. y_{i}^{s}\}_{i=1}^{n_s}$.
Here, $M$ is composed of a feature extractor $f : X \rightarrow \mathbb{R}^d$ and a classifier $g : \mathbb{R}^d \rightarrow \mathbb{R}^K$ where $d$ is the dimension of the feature and $K$ is the number of classes. 
Given the probability $p(x)=\big(p(x)_1, p(x)_2, \cdots, p(x)_K\big)$ where $p(x)_k = p(y=k|x)$, the model-wise probability $p_M(X_t)$ is expressed as follows:
\begin{equation}
    \begin{gathered}
        p_M(X_t) = \textmd{softmax}\big(g(f(X_t))\big), \\
        \textmd{where} \; \textmd{softmax}(z)_c=\cfrac{e^{z_c}}{\Sigma_{c'=1}^K e^{z_{c'}}}.
    \end{gathered}
\label{eq:model_prob}
\end{equation}

\subsection{Preliminary}
We consider a dataset consisting of $n$ i.i.d. samples, $X = \{x_i,y_i,\hat{y}_i\}_{i=1}^n$, where $x_i$ is an input, $y_i$, and $\hat{y}_i$ are the corresponding ground truth label and pseudo-label, respectively.
Following \citet{defineconfidencescore}, we define the confidence score function $\kappa(x_i, \hat{y}_i)$ for ordinal ranking. The confidence score function $\kappa(x_i, \hat{y}_i)$ should return a high score for samples that are more likely to be correctly classified.
\begin{equation}\label{confidence score}
    \begin{gathered}
    \kappa(x_i, \hat{y}_i) \le \kappa(x_j, \hat{y}_j) \Rightarrow \textmd{Pr}[\hat{y}_i=y_i] \le \textmd{Pr}[\hat{y}_j=y_j] \\
    \textmd{ with a high probability of 1}-\delta,    
    \end{gathered}
\end{equation}
where $0\le \delta \le 1$. If $\kappa_1(\cdot)$ is a better confidence score function than $\kappa_2(\cdot)$, then $\delta_1 < \delta_2$.

The most basic and common score is Maxprob which is the maximum value of prediction $p_M(X_t)$. 
Another common score is negative entropy (Ent) which is a negative entropy value of prediction $p_M(X_t)$. 
These two scores only use $p_M(X_t)$; hence, it cannot consider the distribution of the target features.
The Cossim score~\cite{CAN}, which considers data-structure-wise knowledge, uses the cosine similarity between a sample and the center of the cluster that contains the sample based on k-means clustering.
Definition of the listed scores are provided in the Appendix~\ref{Appendix:confidencescores}.

\subsection{Joint Model-Data Structure score}
We propose a novel confidence score, the Joint Model-Data Structure (JMDS) score, which considers both model-wise and data-structure-wise knowledge, unlike existing confidence scores.
The JMDS score consists of two confidence scores: Log-Probability Gap (LPG) and Model Probability of Pseudo-label (MPPL) scores.
Pseudo-labeling based on feature-level clustering is commonly used by other UDA methods~\cite{CAN, srdc} because the decision boundary of the model may violate the cluster assumption.
In this study, GMM is used to cluster the target features and assign pseudo-labels to the target data.
GMM outperforms other clustering methods in terms of confidence measurement because it provides data-structure-wise probability $p_\textmd{data}(X_t)$~\cite{mahalanobis}.
Details of GMM and $p_\textmd{data}(X_t)$ are provided in the Appendix~\ref{Appendix:GMM}.

\paragraph{Data-structure-wise confidence score:}
We propose the Log-Probability Gap (LPG) score as a data-structure-wise confidence score because it uses the log data-structure-wise probability $\log{p_{\textmd{data}}(X_t)}$ obtained from GMM on the target feature space. 
First, we define MINGAP for each sample, the minimum gap from $\log{p_\textmd{data}(x_i^t)_{\hat{y}_i^t}}$ to the other log data-structure-wise probability value.
\begin{equation*}
\begin{gathered}
    \textmd{MINGAP}(x_i^{t}) = \min_a\{\log{p_{\textmd{data}}(x_i^{t})_{\hat{y}_i^t}} - \log{p_{\textmd{data}}(x_i^{t})_a}\}, \\
    \textmd{where} \; \hat{y}_i^t=\argmax_c{p_\textmd{data}(x_i^t)_c}, \; a\in \{1,2,\cdots,K\},\; a\neq\hat{y}_i^t.
\end{gathered}
\end{equation*}
The LPG score is the normalized MINGAP, with a value between [0, 1]. 
\begin{equation}
\label{eq:LPG}
    \begin{gathered}
    \textmd{LPG}(x_i^{t}) = \frac{\textmd{MINGAP}(x_i^{t})}{\max_j{\textmd{MINGAP}(x_j^{t})}}, \\
    \end{gathered}
\end{equation}
where $i,j\in\{1,2,\cdots,n_t\}$. 
It yields high scores for samples that is far from the decision boundary based on GMM.

\paragraph{Model-wise confidence score:}
We propose the Model Probability of Pseudo-label (MPPL) score to include knowledge of the model into the confidence score.
MPPL is the model-wise probability of the corresponding pseudo-label $\hat{Y}_t$.
It provides high scores for samples whose pseudo-label is the same based on $p_M(X_t)$ and $p_\textmd{data}(X_t)$.
\begin{equation}
\label{eq:MPPL}
    \begin{gathered}
        \textmd{MPPL}(x_i^{t}) = p_{M}(x_i^{t})_{\hat{y}_i^{t}}.
    \end{gathered}
\end{equation}

\paragraph{JMDS score:}
An overview of the JMDS score is shown in Figure~\ref{fig:JMDS_overview}.
The JMDS score is the product of LPG and MPPL to emphasize confident samples in both scores, with a value between [0, 1]:
\begin{equation}
\label{eq:JMDS}
    \textmd{JMDS}(x_i^{t}) = \textmd{LPG}(x_i^{t}) \cdot \textmd{MPPL}(x_i^{t}).
\end{equation}
The JMDS score contains knowledge on the data structure from LPG and on the model from MPPL. 
The Maxprob and Ent scores use model prediction only, whereas the Cossim score only uses the data structure. 
In SFUDA, the data structure includes knowledge of the target domain and the model includes knowledge of the source domain.
Therefore, the JMDS score is the only confidence score that considers knowledge from both domains.
The experimental results demonstrating the superiority of the JMDS score over the other scores are presented in Section~\ref{JMDSeval}.

\section{Confidence score Weighting Adaptation using the JMDS}
\label{4_CoWA}

\begin{table*}[t]
    \small
    \caption{Evaluation of the JMDS score based on AURC.}
    \label{Table:JMDSeval}
    \centering
\resizebox{0.92\linewidth}{!}{\begin{tabular}{ccccccccc}
        \toprule
        Dataset & Task & 
        \hspace{-0.25cm} Na\"ive PL \hspace{-0.15cm} + \hspace{-0.15cm} Maxprob \hspace{-0.25cm}
        &
        \hspace{-0.25cm} Na\"ive PL \hspace{-0.15cm} + \hspace{-0.15cm} Ent \hspace{-0.25cm}
        &
        \hspace{-0.25cm} SSPL \hspace{-0.15cm} + \hspace{-0.15cm} Cossim \hspace{-0.25cm}
        &
        \hspace{-0.25cm} GMM \hspace{-0.15cm} + \hspace{-0.15cm} Cossim \hspace{-0.25cm}
        &
        \hspace{-0.25cm} GMM \hspace{-0.15cm} + \hspace{-0.15cm} MPPL \hspace{-0.25cm}
        &
        \hspace{-0.25cm} GMM \hspace{-0.15cm} + \hspace{-0.15cm} LPG \hspace{-0.25cm}
        &
        \hspace{-0.25cm} GMM \hspace{-0.15cm} + \hspace{-0.15cm} JMDS \hspace{-0.25cm} \\
                                      \midrule
\multirow{7}{*}{Office-31}

 &  A  $\rightarrow$  D  & 0.047 & 0.051 & \textbf{0.018} & 0.031 & 0.039 & 0.033 & 0.033 \\
 &  A  $\rightarrow$  W  & 0.074 & 0.081 & \textbf{0.034} & 0.045 & 0.059 & 0.042 & 0.044 \\
 &  D  $\rightarrow$  A  & 0.158 & 0.165 & 0.140 & 0.130 & 0.131 & 0.127 & \textbf{0.115} \\
 &  D  $\rightarrow$  W  & 0.007 & 0.008 & 0.009 & 0.009 & 0.005 & 0.004 & \textbf{0.004} \\
 &  W  $\rightarrow$  A  & 0.157 & 0.167 & \textbf{0.107} & 0.108 & 0.132 & 0.120 & 0.113 \\
 &  W  $\rightarrow$  D  & 0.002 & 0.002 & \textbf{0.001} & \textbf{0.001} & \textbf{0.001} & \textbf{0.001} & \textbf{0.001} \\
 \cmidrule{2-9}
 & Avg. & 0.074 & 0.079 & \textbf{0.052} & 0.054 & 0.061 & 0.055 & \textbf{0.052} \\
 
                                      \midrule
\multirow{13}{*}{Office-Home}
 &  Ar  $\rightarrow$  Cl  & 0.308 & 0.316 & 0.296 & 0.274 & 0.278 & 0.265 & \textbf{0.256} \\
 &  Ar  $\rightarrow$  Pr  & 0.140 & 0.145 & \textbf{0.100} & 0.105 & 0.116 & 0.125 & 0.104 \\
 &  Ar  $\rightarrow$  Rw  & 0.088 & 0.095 & 0.086 & 0.086 & 0.076 & 0.086 & \textbf{0.068} \\
 &  Cl  $\rightarrow$  Ar  & 0.238 & 0.249 & 0.200 & 0.194 & 0.212 & 0.216 & \textbf{0.197} \\
 &  Cl  $\rightarrow$  Pr  & 0.159 & 0.168 & \textbf{0.105} & 0.113 & 0.131 & 0.125 & 0.115 \\
 &  Cl  $\rightarrow$  Rw  & 0.151 & 0.159 & 0.113 & 0.113 & 0.125 & 0.115 & \textbf{0.106} \\
 &  Pr  $\rightarrow$  Ar  & 0.237 & 0.246 & 0.185 & \textbf{0.184} & 0.210 & 0.214 & 0.190 \\
 &  Pr  $\rightarrow$  Cl  & 0.365 & 0.375 & 0.339 & 0.315 & 0.327 & 0.293 & \textbf{0.293} \\
 &  Pr  $\rightarrow$  Rw  & 0.095 & 0.099 & 0.080 & 0.082 & 0.084 & 0.091 & \textbf{0.073} \\
 &  Rw  $\rightarrow$  Ar  & 0.138 & 0.147 & 0.129 & 0.125 & 0.126 & 0.154 & \textbf{0.118} \\
 &  Rw  $\rightarrow$  Cl  & 0.314 & 0.325 & 0.298 & 0.284 & 0.275 & 0.248 & \textbf{0.238} \\
 &  Rw  $\rightarrow$  Pr  & 0.073 & 0.078 & 0.062 & 0.063 & 0.065 & 0.078 & \textbf{0.059} \\
 \cmidrule{2-9}
 & Avg. & 0.192 & 0.200 & 0.166 & 0.162 & 0.169 & 0.168 & \textbf{0.151} \\
                                      \midrule              
\multirow{1}{*}{VisDA-2017}
 & T $\rightarrow$ V 
 & 0.274 & 0.284 & 0.261 & 0.202 & 0.204 & 0.172 & \textbf{0.162} \\
        \bottomrule                    
\end{tabular}}
\vspace{-0.25cm}
\end{table*}

We aim to produce high performance on the target domain by fine-tuning model $M$ using pseudo-labels $\hat{Y}_t$ from GMM and the JMDS score.
A simple way to exploit the confidence score is sample weighting which is effective for robust learning such as learning with noisy labels~\cite{reweightrobust}. 
SFUDA basically includes incorrect pseudo-labels; hence, we expect the sample weighting using confidence scores to be effective for SFUDA.
Therefore, we propose a Confidence Score Weighting Adaptation using the JMDS (CoWA-JMDS) framework whose loss is as follows:
\begin{equation}
    \begin{gathered}
        \mathcal{L}_{\textmd{CoWA-JMDS}}({x}_i^{t}) = {\textmd{JMDS}}(x_i^t)\cdot\mathcal{L}_{\textmd{CE}}({p_{M}({x}_i^{t}), \hat{y}_i^t}),
    \end{gathered}
\label{eq:CoWA-JMDS}
\end{equation}
where $\mathcal{L}_{\textmd{CE}}({p_{M}({x}_i^{t}), \hat{y}_i^t})=-\log{p_{M}(x_i^t)_{\hat{y}_i^t}}$ is the cross entropy loss.
The pseudo-code for CoWA-JMDS is provided in the Appendix~\ref{Appendix:pseudocode}.

\paragraph{Weight Mixup:}
CoWA-JMDS rarely allows low-confidence samples whose JMDS scores close to 0 to participate in training so that the model can learn robustly with incorrect pseudo-labels. 
However, this means that the knowledge provided by the target feature distribution is not fully utilized. 
Therefore, we propose a technique called weight Mixup, a variant of Mixup~\cite{mixup}, to utilize more knowledge of the target feature distribution.

Mixup mixes images and corresponding labels. All mixed samples had the same sample-wise weights for training. However, in CoWA-JMDS, the sample has its own confidence score as a sample-wise weight for training. This means that mixed images that use low-confidence samples for mixing should have a lower sample-wise weight for Mixup training.
Therefore, the proposed weight Mixup mixes the corresponding sample-wise weights together.
\begin{equation}
\label{eq:mixup}
    \begin{gathered}
        \tilde{x}^{t} = \gamma \cdot x_i^{t} + (1 - \gamma) \cdot x_j^{t}, \\
        \tilde{y}^{t} = \gamma \cdot o(\hat{y}_i^{t}) + (1 - \gamma) \cdot o(\hat{y}_j^{t}), \\
        w(\tilde{x}^{t}) = \gamma \cdot \textmd{JMDS}(x_i^{t}) + (1 - \gamma) \cdot \textmd{JMDS}(x_j^{t}),\\
    \end{gathered}
\end{equation}
where $\gamma \sim \textmd{Beta}(\alpha, \alpha)$, for $\alpha \in (0, \infty)$, and $o(\cdot)$ is a one-hot encoding function. The loss function of the weight Mixup is modified as follows:
\begin{equation}
    \begin{gathered}
        \mathcal{L}_{\textmd{Mixup}}(\tilde{x}^{t}, \tilde{y}^{t}) = w(\tilde{x}^{t}) \cdot \mathbb{E}_{\tilde{y}^{t}}[-\log{p_{{M}}(\tilde{x}^{t})}]
    \end{gathered}
    \label{eq:CoWA-JMDS*}
\end{equation}
Weight Mixup makes the training robust to incorrect pseudo-labels in the following manner: 
A mixture of low- and high-confidence samples will produce a sample with mid-level confidence, which can robustly and effectively participate in the learning. By contrast, the resulting sample will be suppressed when low-confidence samples are mixed with each other. 
Therefore, weight Mixup is helpful for CoWA-JMDS as it enhances target feature knowledge by augmenting samples with the mid-level confidence.

\subsection{General UDA scenarios}
CoWA-JMDS can be easily extended to more general UDA settings.
Open-set~\cite{oda} and partial-set~\cite{pda} scenarios are relatively realistic scenarios where only a few categories of interest are shared between the source and target data.

In an open-set scenario, the target domain contains unseen classes that are not included in the source domain. 
Therefore, we should classify known classes included in the source domain and unknown classes. Algorithm~\ref{alg:oda} in the Appendix~\ref{Appendix:algorithms} shows the process of classifying the known and unknown classes.
Following the protocols in \citet{shot}, we sorted the entropy of the samples and performed a two-class k-means clustering. Then, a high-entropy cluster was classified as unknown samples, whereas a low-entropy cluster was classified as known samples. Known samples were used to train the models.

\begin{table}[t]
    \small
    \vspace{-0.25cm}
    \caption{Accuracy (\%) on Office-31 dataset for UDA and SFUDA methods (ResNet-50).}
    \label{office_result}
    \centering
    \resizebox{\linewidth}{!}
    {\begin{tabular}{clccccccc}
        \toprule
        Task                    & Method          & A$\rightarrow$D & A$\rightarrow$W & D$\rightarrow$A & D$\rightarrow$W & W$\rightarrow$A & W$\rightarrow$D & Avg. \\
        \midrule
        \multirow{6}{*}{SFUDA} 
        & SFIT~\citep{sfit}          & 89.9              & 91.8              & 73.9              & \underline{98.7}              & 72.0              & \underline{99.9}              & 87.7 \\
        & SHOT~\cite{shot}            & 94.0              & 90.1              & 74.7              & 98.4              & 74.3              & \underline{99.9}              & 88.6 \\
                                & 3C-GAN~\cite{3cgan}          & 92.7              & \underline{93.7}              & 75.3              & 98.5              & \textbf{77.8}              & 99.8              & \underline{89.6} \\
                                & NRC~\cite{nrc}          & \textbf{96.0}              & 90.8              &  75.3             & \textbf{99.0}             &    75.0           &    \textbf{100.0}           & 89.4  \\
                                & CoWA-JMDS (w/o weight Mixup)          & 93.7                  &    93.5           &    \underline{75.5}           &    98.0           &    76.8           &    99.8           & \underline{89.6} \\
                                & CoWA-JMDS          & \underline{94.4}              & \textbf{95.2}              &  \textbf{76.2}             & 98.5             &    \underline{77.6}           &    99.8           & \textbf{90.3}  \\
        \midrule
        \multirow{4}{*}{UDA}    & ResNet~\cite{resnet}          & 68.9              & 68.4              & 62.5              & 96.7              & 60.7              & 99.3              & 76.1 \\
                                & CAN~\cite{CAN}             & {95.0}              & 94.5              & {78.0}              & {99.1}              & 77.0              & 99.8              & 90.6 \\
                                & RSDA-MSTN~\cite{RSDA-MSTN}       & {95.8}              & {96.1}              & 77.4              & {99.3}              & {78.9}              & {100}               & {91.1} \\
                                & FixBi~\cite{fixbi}           & {95.0}              & {96.1}              & {78.7}              & {99.3}              & {79.4}              & {100}               & {91.4} \\
        \bottomrule                        
    \end{tabular}}
    \vspace{-0.55cm}
\end{table}

In a partial-set scenario, the target domain contains a few classes included in the source domain.
Therefore, the absent classes should be filtered out.
Algorithm~\ref{alg:pda} in the Appendix~\ref{Appendix:algorithms} shows how to estimate the classes included in the target domain.
First, we initialized the parameters for GMM using the prediction of the model and perform the GMM expectation-maximization iteration once.
Next, we filtered out classes whose number of samples were lower than a threshold.
Finally, the aforementioned two steps were iteratively executed until there were no filtered out classes.

\section{Experiments}
\label{5_Experiments}
We evaluated our proposed methods, the JMDS score and CoWA-JMDS, on three public UDA benchmarks: Office-31~\cite{office31}, Office-Home~\cite{officehome}, and VisDA-2017~\cite{visda}. 
More details are provided in the Appendix~\ref{Appendix:impdetail}.
\vspace{-0.1cm}
\begin{table*}[t]
  \small
  \caption{Accuracy (\%) on Office-Home for UDA and source-free UDA methods (ResNet-50).}
  \label{office_home_result}
  \centering
  \resizebox{0.95\linewidth}{!}{\begin{tabular}{clccccccccccccc}
    \toprule
    Task
    & Method 
    & \hspace{-0.25cm} Ar \hspace{-0.15cm} $\rightarrow$ \hspace{-0.15cm} Cl \hspace{-0.25cm}
    & \hspace{-0.25cm} Ar \hspace{-0.15cm} $\rightarrow$ \hspace{-0.15cm} Pr \hspace{-0.25cm}
    & \hspace{-0.25cm} Ar \hspace{-0.15cm} $\rightarrow$ \hspace{-0.15cm} Rw \hspace{-0.25cm}
    & \hspace{-0.25cm} Cl \hspace{-0.15cm} $\rightarrow$ \hspace{-0.15cm} Ar \hspace{-0.25cm}
    & \hspace{-0.25cm} Cl \hspace{-0.15cm} $\rightarrow$ \hspace{-0.15cm} Pr \hspace{-0.25cm}
    & \hspace{-0.25cm} Cl \hspace{-0.15cm} $\rightarrow$ \hspace{-0.15cm} Rw \hspace{-0.25cm}
    & \hspace{-0.25cm} Pr \hspace{-0.15cm} $\rightarrow$ \hspace{-0.15cm} Ar \hspace{-0.25cm}
    & \hspace{-0.25cm} Pr \hspace{-0.15cm} $\rightarrow$ \hspace{-0.15cm} Cl \hspace{-0.25cm}
    & \hspace{-0.25cm} Pr \hspace{-0.15cm} $\rightarrow$ \hspace{-0.15cm} Rw \hspace{-0.25cm}
    & \hspace{-0.25cm} Rw \hspace{-0.15cm} $\rightarrow$ \hspace{-0.15cm} Ar \hspace{-0.25cm}
    & \hspace{-0.25cm} Rw \hspace{-0.15cm} $\rightarrow$ \hspace{-0.15cm} Cl \hspace{-0.25cm}
    & \hspace{-0.25cm} Rw \hspace{-0.15cm} $\rightarrow$ \hspace{-0.15cm} Pr \hspace{-0.25cm}
    & Avg
    \\
    \midrule
    \multirow{5}{*}{SFUDA}
    & BAIT~\citep{bait}             
    & \underline{57.4} & {77.5} & \textbf{82.4} & 68.0 & 77.2 & 75.1 & 67.1 & 55.5 & {81.9} & \textbf{73.9} & \underline{59.5} & 84.2 & 71.6 \\
    & SHOT~\cite{shot}             
    & 57.1 & {78.1} & {81.5} & 68.0 & 78.2 & 78.1 & 67.4 & 54.9 & {82.2} & 73.3 & 58.8 & 84.3 & 71.8 \\
    & NRC~\cite{nrc} 
    & \textbf{57.7} & \textbf{80.3} & \underline{82.0} & 68.1 & \underline{79.8} & 78.6 & 65.3 & 56.4 & \textbf{83.0} & 71.0 & 58.6 & \textbf{85.6} & \underline{72.2} \\
    & CoWA-JMDS (w/o weight Mixup) 
    & 56.4 & \underline{78.6} & 80.3 & \underline{68.8} & 79.7 & \underline{78.7} & \textbf{68.1} & \underline{56.8} & 82.0 & \underline{73.4} & 59.1 & 83.9 & \underline{72.2} \\
    & CoWA-JMDS 
    & 56.9 & 78.4 & 81.0 & \textbf{69.1} & \textbf{80.0} & \textbf{79.9} & \underline{67.7} & \textbf{57.2} & \underline{82.4} & 72.8 & \textbf{60.5} & \underline{84.5} & \textbf{72.5} \\
    \midrule
    \multirow{3}{*}{UDA}
    & ResNet-50~\cite{resnet}
    & 34.9 & 50.0 & 58.0 & 37.4 & 41.9 & 46.2 & 38.5 & 31.2 & 60.4 & 53.9 & 41.2 & 59.9 & 46.1 \\
    & RSDA-MSTN~\cite{RSDA-MSTN} \hspace{-0.25cm}
    & 53.2 & 77.7 & 81.3 & 66.4 & 74.0 & 76.5 & {67.9} & 53.0 & {82.0} & {75.8} & 57.8 & {85.4} & 70.9 \\
    & FixBi~\cite{fixbi}             
    & 58.1 & 77.3 & 80.4 & 67.7 & 79.5 & 78.1 & 65.8 & {57.9} & 81.7 & {76.4} & {62.9} & {86.7} & 72.7 \\
    \bottomrule
  \end{tabular}}
  \vspace{-0.1cm}
\end{table*}
\begin{table*}[t]
  \vspace{-0.45 cm}
  \small
  \caption{Accuracy (\%) on VisDA-2017 for UDA and source-free UDA methods (ResNet-101).}
  \label{visda_result}
  \centering
  \resizebox{0.95\linewidth}{!}{\begin{tabular}{clccccccccccccc}
    \toprule
    Task
    & Method 
    & {plane} 
    & {bcycl} 
    & {bus} 
    & {car} 
    & {horse} 
    & {knife} 
    & {mcycl} 
    & {person} 
    & {plant} 
    & {sktbrd} 
    & {train} 
    & {truck} 
    & Average 
    \\
    \midrule
    \multirow{6}{*}{SFUDA}
    & SFIT~\citep{sfit} 
    & 94.3 & {79.0} & \textbf{84.9} & 63.6 & 92.6 & 92.0 & 88.4 & 79.1 & 92.2 & 79.8 & 87.6 & 43.0 & 81.4 \\
    & 3C-GAN~\cite{3cgan}
    & 94.8 & 73.4 & 68.8 & \textbf{74.8} & 93.1 & {95.4} & \underline{88.6} & \underline{84.7} & 89.1 & 84.7 & 83.5 & 48.1 & 81.6 \\
    & SHOT~\cite{shot} 
    & 94.3 & {88.5} & 80.1 & 57.3 & 93.1 & 94.9 & 80.7 & 80.3 & 91.5 & 89.1 & 86.3 & \underline{58.2} & 82.9 \\
    & NRC~\cite{nrc} 
    & \textbf{96.8} & \textbf{91.3} & 82.4 & 62.4 & \underline{96.2} & \underline{95.9} & 86.1 & 80.6 & \textbf{94.8} & \textbf{94.1} & \textbf{90.4} & \textbf{59.7} & \underline{85.9} \\
    & CoWA-JMDS (w/o weight Mixup)
    & \underline{96.3} & 88.5 & \underline{84.1} & 59.7 & 95.2 & \underline{96.9} & 82.1 & 82.3 & 93.3 & \underline{92.8} & 87.5 & 51.1 & 84.2 \\

    & CoWA-JMDS
    & \underline{96.2} & \underline{89.7} & \textbf{83.9} & \underline{73.8} & \textbf{96.4} & \textbf{97.4} & \textbf{89.3} & \textbf{86.8} & \underline{94.6} & \underline{92.1} & \underline{88.7} & 53.8 & \textbf{86.9} \\

    \midrule
    \multirow{3}{*}{UDA}
    & ResNet-101~\cite{resnet} & 72.3 & 6.1 & 63.4 & {91.7} & 52.7 & 7.9 & 80.1 & 5.6 & 90.1 & 18.5 & 78.1 & 25.9 & 49.4 \\
    & CAN~\cite{CAN}      & {97.0} & 87.2 & 82.5 & 74.3 & {97.8} & 96.2 & {90.8} & 80.7 & {96.6} & {96.3} & 87.5 & 59.9 & 87.2 \\
    & FixBi~\cite{fixbi}             & 96.1 & {87.8} & {{90.5}} & {{90.3}} & {96.8} & 95.3 & {{92.8}} & {88.7} & {97.2} & 94.2 & {90.9} & 25.7 & {87.2} \\
    \bottomrule
  \end{tabular}}
  \vspace{-0.35cm}
\end{table*}
\begin{table*}[t]
  \small
  \caption{Accuracy (\%) on Office-Home for open-set and partial-set scenarios (ResNet-50).}
  \label{office_home_oda_pda}
  \centering
  \resizebox{0.95\linewidth}{!}{\begin{tabular}{clccccccccccccc}
    \toprule
    Task (Open-set)
    & Method 
    & \hspace{-0.25cm} Ar \hspace{-0.15cm} $\rightarrow$ \hspace{-0.15cm} Cl \hspace{-0.25cm}
    & \hspace{-0.25cm} Ar \hspace{-0.15cm} $\rightarrow$ \hspace{-0.15cm} Pr \hspace{-0.25cm}
    & \hspace{-0.25cm} Ar \hspace{-0.15cm} $\rightarrow$ \hspace{-0.15cm} Rw \hspace{-0.25cm}
    & \hspace{-0.25cm} Cl \hspace{-0.15cm} $\rightarrow$ \hspace{-0.15cm} Ar \hspace{-0.25cm}
    & \hspace{-0.25cm} Cl \hspace{-0.15cm} $\rightarrow$ \hspace{-0.15cm} Pr \hspace{-0.25cm}
    & \hspace{-0.25cm} Cl \hspace{-0.15cm} $\rightarrow$ \hspace{-0.15cm} Rw \hspace{-0.25cm}
    & \hspace{-0.25cm} Pr \hspace{-0.15cm} $\rightarrow$ \hspace{-0.15cm} Ar \hspace{-0.25cm}
    & \hspace{-0.25cm} Pr \hspace{-0.15cm} $\rightarrow$ \hspace{-0.15cm} Cl \hspace{-0.25cm}
    & \hspace{-0.25cm} Pr \hspace{-0.15cm} $\rightarrow$ \hspace{-0.15cm} Rw \hspace{-0.25cm}
    & \hspace{-0.25cm} Rw \hspace{-0.15cm} $\rightarrow$ \hspace{-0.15cm} Ar \hspace{-0.25cm}
    & \hspace{-0.25cm} Rw \hspace{-0.15cm} $\rightarrow$ \hspace{-0.15cm} Cl \hspace{-0.25cm}
    & \hspace{-0.25cm} Rw \hspace{-0.15cm} $\rightarrow$ \hspace{-0.15cm} Pr \hspace{-0.25cm}
    & Avg
    \\
    \midrule
    \multirow{3}{*}{SFUDA}
    & SHOT~\citep{shot}             
    & \underline{64.5} & \textbf{80.4} & {84.7} & 63.1 & {75.4} & 81.2 & \underline{65.3} & \textbf{59.3} & \underline{83.3} & \underline{69.6} & \textbf{64.6} & 82.3 & 72.8 \\
    & CoWA-JMDS (w/o weight Mixup)
    & \textbf{64.6} & \underline{80.2} & \textbf{88.1} & \underline{67.3} & \underline{83.5} & \textbf{82.2} & 63.9 & \underline{57.1} & \textbf{84.4} & \textbf{70.8} & \underline{64.0} & \underline{84.8} & \textbf{74.2} \\
    & CoWA-JMDS
    & 63.3 & 79.2 & \underline{85.4} & \textbf{67.6} & \textbf{83.6} & \underline{82.0} & \textbf{66.9} & 56.9 & 81.1 & 68.5 & 57.9 & \textbf{85.9} & \underline{73.2}
 \\
    \midrule
    \multirow{3}{*}{UDA}
    & ResNet-50~\citep{resnet}
    & 53.4 & 52.7 & 51.9 & 69.3 & 61.8 & 74.1 & 61.4 & 64.0 & 70.0 & 78.7 & 71.0 & 74.9 & 64.3 \\
    & STA~\citep{sta} \hspace{-0.25cm}
    & 58.1 & 53.1 & 54.4 & 71.6 & 69.3 & 81.9 & 63.4 & 65.2 &  74.9 & 85.0 & 75.8 & 80.8 & 69.5 \\
    & PGL~\citep{pgl}             
    & 61.6 & 77.1 & 85.9 & 68.8 & 72.0 & 82.8 & 72.2 & 58.4 & 82.6 & 78.6 &  65.0 &  83.0 & 74.0 \\
    \bottomrule
    \toprule
    Task (Partial-set)
    & Method 
    & \hspace{-0.25cm} Ar \hspace{-0.15cm} $\rightarrow$ \hspace{-0.15cm} Cl \hspace{-0.25cm}
    & \hspace{-0.25cm} Ar \hspace{-0.15cm} $\rightarrow$ \hspace{-0.15cm} Pr \hspace{-0.25cm}
    & \hspace{-0.25cm} Ar \hspace{-0.15cm} $\rightarrow$ \hspace{-0.15cm} Rw \hspace{-0.25cm}
    & \hspace{-0.25cm} Cl \hspace{-0.15cm} $\rightarrow$ \hspace{-0.15cm} Ar \hspace{-0.25cm}
    & \hspace{-0.25cm} Cl \hspace{-0.15cm} $\rightarrow$ \hspace{-0.15cm} Pr \hspace{-0.25cm}
    & \hspace{-0.25cm} Cl \hspace{-0.15cm} $\rightarrow$ \hspace{-0.15cm} Rw \hspace{-0.25cm}
    & \hspace{-0.25cm} Pr \hspace{-0.15cm} $\rightarrow$ \hspace{-0.15cm} Ar \hspace{-0.25cm}
    & \hspace{-0.25cm} Pr \hspace{-0.15cm} $\rightarrow$ \hspace{-0.15cm} Cl \hspace{-0.25cm}
    & \hspace{-0.25cm} Pr \hspace{-0.15cm} $\rightarrow$ \hspace{-0.15cm} Rw \hspace{-0.25cm}
    & \hspace{-0.25cm} Rw \hspace{-0.15cm} $\rightarrow$ \hspace{-0.15cm} Ar \hspace{-0.25cm}
    & \hspace{-0.25cm} Rw \hspace{-0.15cm} $\rightarrow$ \hspace{-0.15cm} Cl \hspace{-0.25cm}
    & \hspace{-0.25cm} Rw \hspace{-0.15cm} $\rightarrow$ \hspace{-0.15cm} Pr \hspace{-0.25cm}
    & Avg
    \\
    \midrule
    \multirow{3}{*}{SFUDA}
    & SHOT~\citep{shot}             
    & 64.8 & 85.2 & \textbf{92.7} & \textbf{76.3} & 77.6 & 88.8 & 79.7 & 64.3 & 89.5 & 80.6 & 66.4 & 85.8 & 79.3 \\
    & CoWA-JMDS (w/o weight Mixup)
    & \textbf{69.7} & \underline{91.6} & 92.1 & \textbf{78.9} & \textbf{86.3} & \underline{91.6} & \textbf{81.5} & \underline{64.4} & \underline{89.7} & \textbf{84.1} & \underline{71.6} & \underline{90.2} & \underline{82.6} \\
    & CoWA-JMDS
    & \underline{69.6} & \textbf{93.2} & \underline{92.3} & \textbf{78.9} & \underline{81.3} & \textbf{92.1} & \underline{79.8} & \textbf{71.7} & \textbf{90.0} & \underline{83.8} & \textbf{72.2} & \textbf{93.7} & \textbf{83.2} \\
    \midrule
    \multirow{3}{*}{UDA}
    & ResNet-50~\citep{resnet}
    & 46.3 & 67.5 & 75.9 & 59.1 & 59.9 & 62.7 & 58.2 & 41.8 & 74.9 & 67.4 & 48.2 & 74.2 & 61.3 \\
    & SAFN~\citep{safn} \hspace{-0.25cm}
    & 58.9 & 76.3 & 81.4 & 70.4 & 73.0 & 77.8 & {72.4} & 55.3 &  {80.4} & {75.8} & 60.4 & {79.9} & 71.8 \\
    & $\textmd{BA}^{3}$US~\citep{baus}             
    & 60.6 & 83.1 & 88.4 & 71.8 & 72.8 & 83.4 & 75.5 & 61.6 & 86.5 & 79.3 &  62.8 &  {86.1} & 76.0 \\
    \bottomrule
  \end{tabular}}
  \vspace{-0.3cm}
\end{table*}

\subsection{JMDS evaluation}
\label{JMDSeval}
We introduced the JMDS score as a reliable confidence score. 
To prove its efficacy, we compared the JMDS score with other scores~\cite{selective3, CAN}.
The pseudo-label of Maxprob and Ent scores is given by the index of maximum model probability $\textmd{argmax}_{c} \, {p_M(x_i^{t})}_c$ following naive PL~\cite{pseudolabel}. 
SSPL, proposed by \citet{shot}, and GMM provide the pseudo-label for Cossim.
SSPL uses modified k-means clustering to generate pseudo-labels.

To evaluate the confidence score, we measured the Area Under Risk-Coverage curve (AURC)~\cite{revisitingconfidencescore} using the 0/1 loss function, which returns a value of one if the pseudo- and ground truth-labels of the sample are different, and a value of zero if they are the same. 
\citet{revisitingconfidencescore} compared the Area Under Receiver Operating Characteristic curve (AUROC), Area Under Precision-Recall curve (AUPR), and AURC. 
They claimed that AURC is the only reliable measurement when the underlying models are the same. 
The risk-coverage curve was first proposed by \citet{defineconfidencescore}. 
After obtaining the high-confidence set, $X^h_t = \{x_i^{t} | \kappa(x_i^{t}, \hat{y}_i^{t}) > \tau\}$, where $\tau$ is a threshold, risk is the average empirical loss of $X^h_t$, and the coverage is $|X^h_t| / |X_t|$.
A lower AURC value indicates higher reliability because it implies a lower risk for the same coverage. When 0/1 loss is applied, a high AURC indicates low correctness for corresponding pseudo-labels.

The experimental results are presented in Table~\ref{Table:JMDSeval}. 
We measured the reliability of various confidence scores for the pre-trained source model obtained using five random seeds.
The best AURC is indicated in bold.
We quantitatively compared various confidence measuring strategies based on AURC values.
The JMDS score achieved the lowest AURC value in most adaptation tasks.
Using the MPPL or LPG score alone is worse than using the JMDS score, which considers both scores. 
This demonstrates that using knowledge of the model and data structure jointly is superior to considering only one aspect.

\subsection{CoWA-JMDS evaluation}

We evaluated model $M$ trained by CoWA-JMDS and compared it with SFUDA~\citep{shot, 3cgan, bait, sfit, nrc} and conventional UDA baseline methods ~\citep{CAN, RSDA-MSTN, fixbi}. Notably, our task is SFUDA, which is a more challenging task than conventional UDA that directly uses the source data during training.
\textcolor{black}{We trained the models using five different random seeds and reported their average performance.}
The best accuracy is indicated in bold, and the second-best accuracy is underlined. 

\paragraph{Closed-set scenario:}
Table~\ref{office_result},~\ref{office_home_result}, and~\ref{visda_result} show the classification accuracy for a closed-set scenario in all tasks on each dataset: Office-31, Office-Home, and VisDA-2017. 
CoWA-JMDS achieved the best performance for all three datasets. CoWA-JMDS boosts the best performance 0.7\% on the Office-31 dataset, 0.3\% on the Office-Home dataset, and 1.0\% on the VisDA-2017 dataset.
The results demonstrate that our proposed CoWA-JMDS framework is effective for SFUDA.
Even, despite its simplicity, we obtained the same performance with the state-of-the-art method on the Office-31, and Office-Home dataset without weight Mixup, only using the JMDS score as a sample weight and the cross entropy loss.

\paragraph{Extended UDA scenarios:}
We evaluated CoWA-JMDS for open-set and partial-set scenarios on the Office-Home dataset.
Following the protocols in ~\citet{shot}, the source domain consists of 25 classes (the first 25 in alphabet order) but the target domain contains 65 classes including unknown samples for an open-set scenario.
However, for a partial-set scenario, the source domain consists of 65 classes but the target domain contains the same 25 classes.

Table~\ref{office_home_oda_pda} shows the results of experiments for the open-set and partial-set scenarios.
CoWA-JMDS achieved the best performance among all methods including conventional UDA methods for both scenarios.
In an open-set scenario, CoWA-JMDS without weight Mixup outperforms CoWA-JMDS with weight Mixup because weight Mixup had a negative effect when the known-classified unknown class samples were mixed with other samples.

\section{Further analysis}
\label{6_Discussion}
\begin{figure*}[t]
    \centering
    \subfigure[\label{fig:source_dist}]{\includegraphics[width=.188\linewidth]{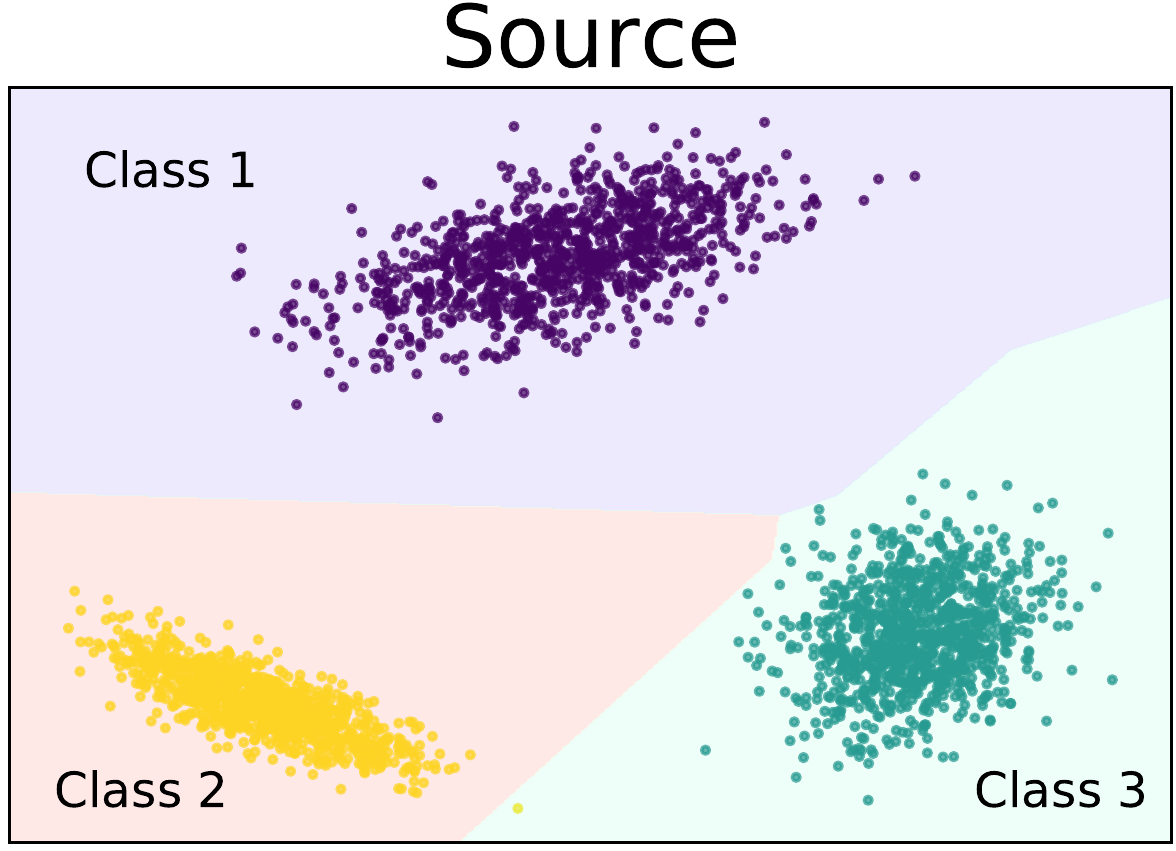}}
    \vspace{-0.275cm}
    \subfigure[\label{fig:conf_maxprob}]{\includegraphics[width=.18\linewidth]{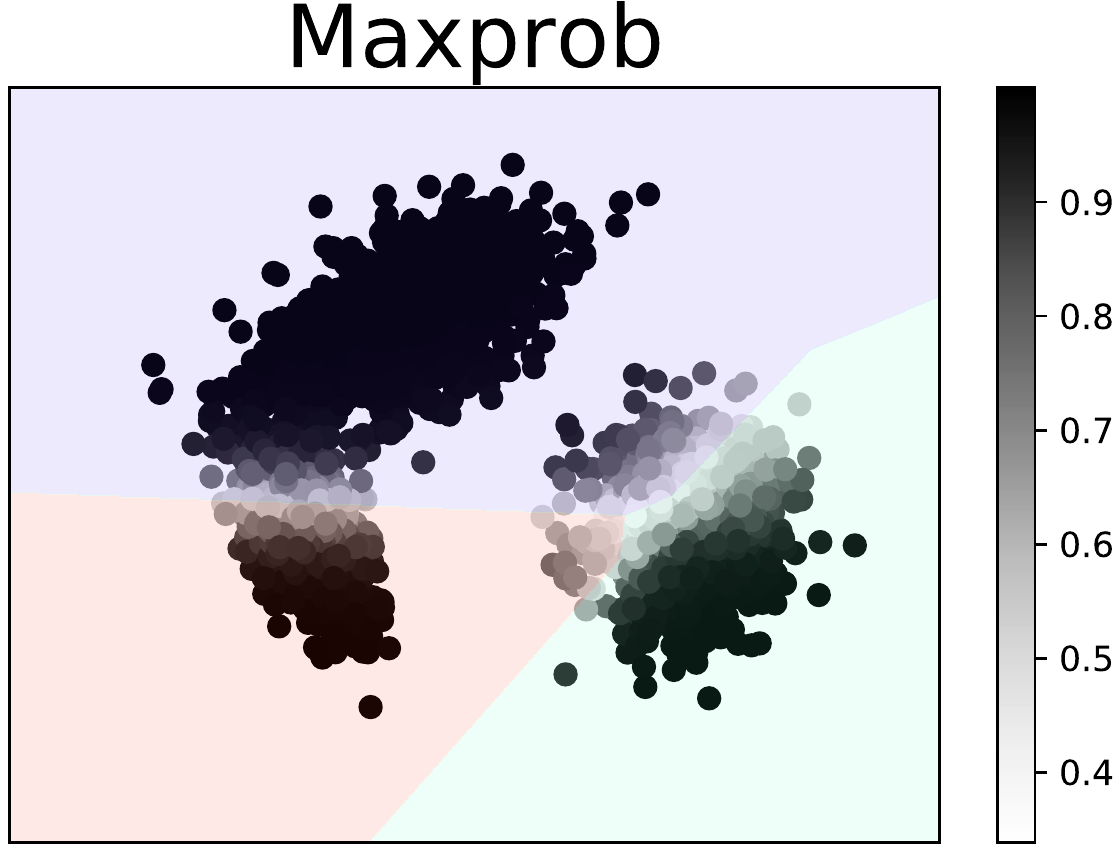}}
    \subfigure[\label{fig:conf_mppl_data}]{\includegraphics[width=.18\linewidth]{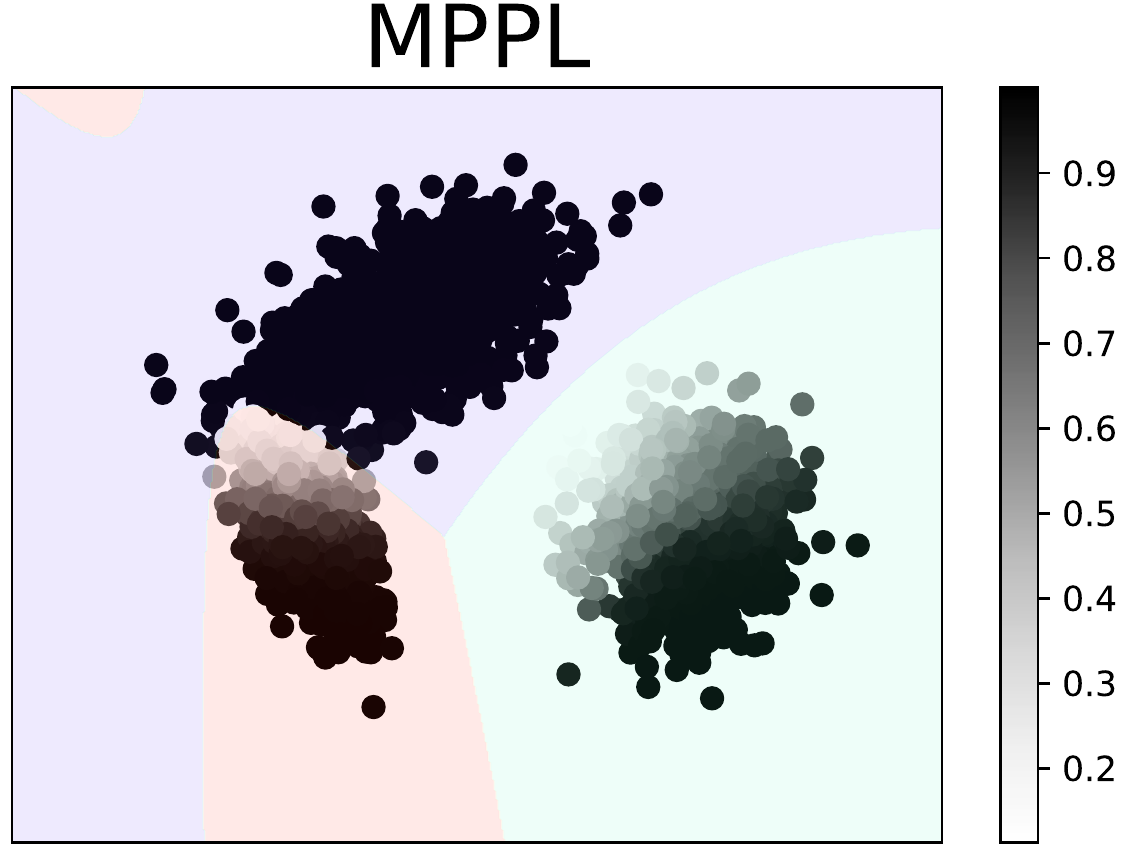}}
    \subfigure[\label{fig:conf_lpg_data}]{\includegraphics[width=.18\linewidth]{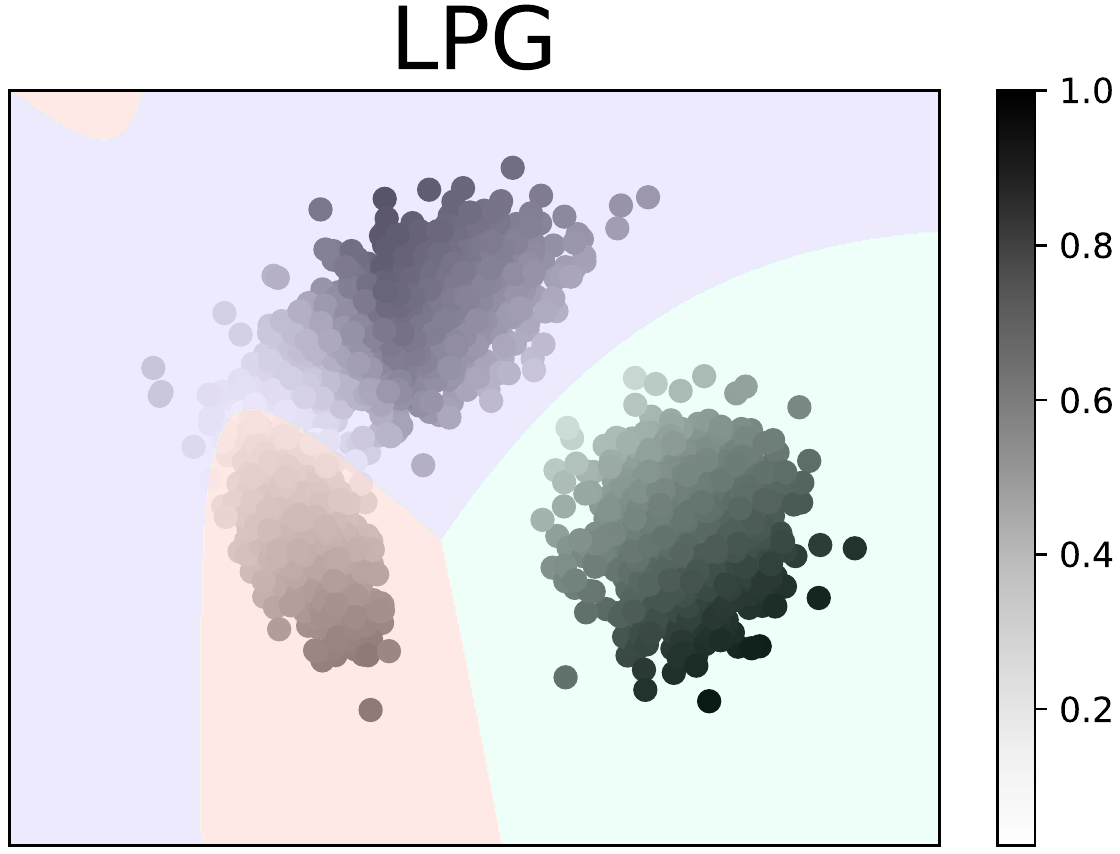}}
    \subfigure[\label{fig:conf_jmds_data}]{\includegraphics[width=.18\linewidth]{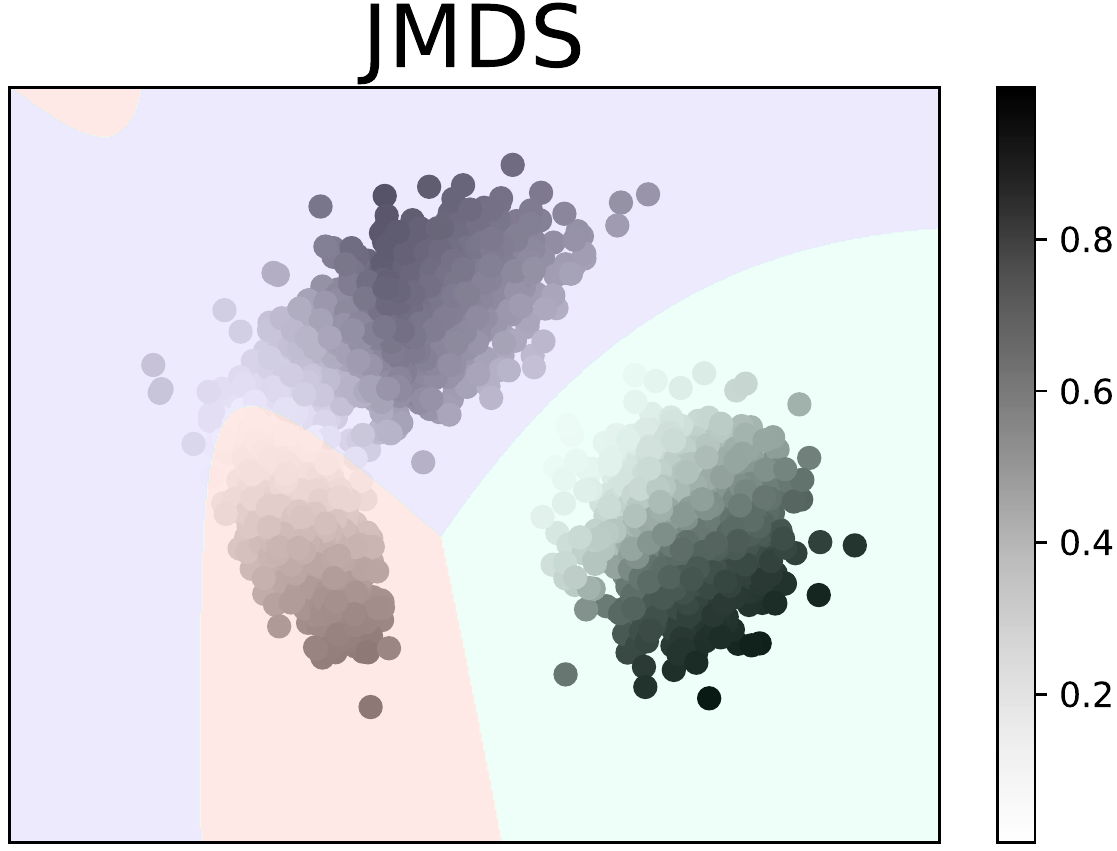}}
    \subfigure[\label{fig:target_dist}]{\includegraphics[width=.188\linewidth]{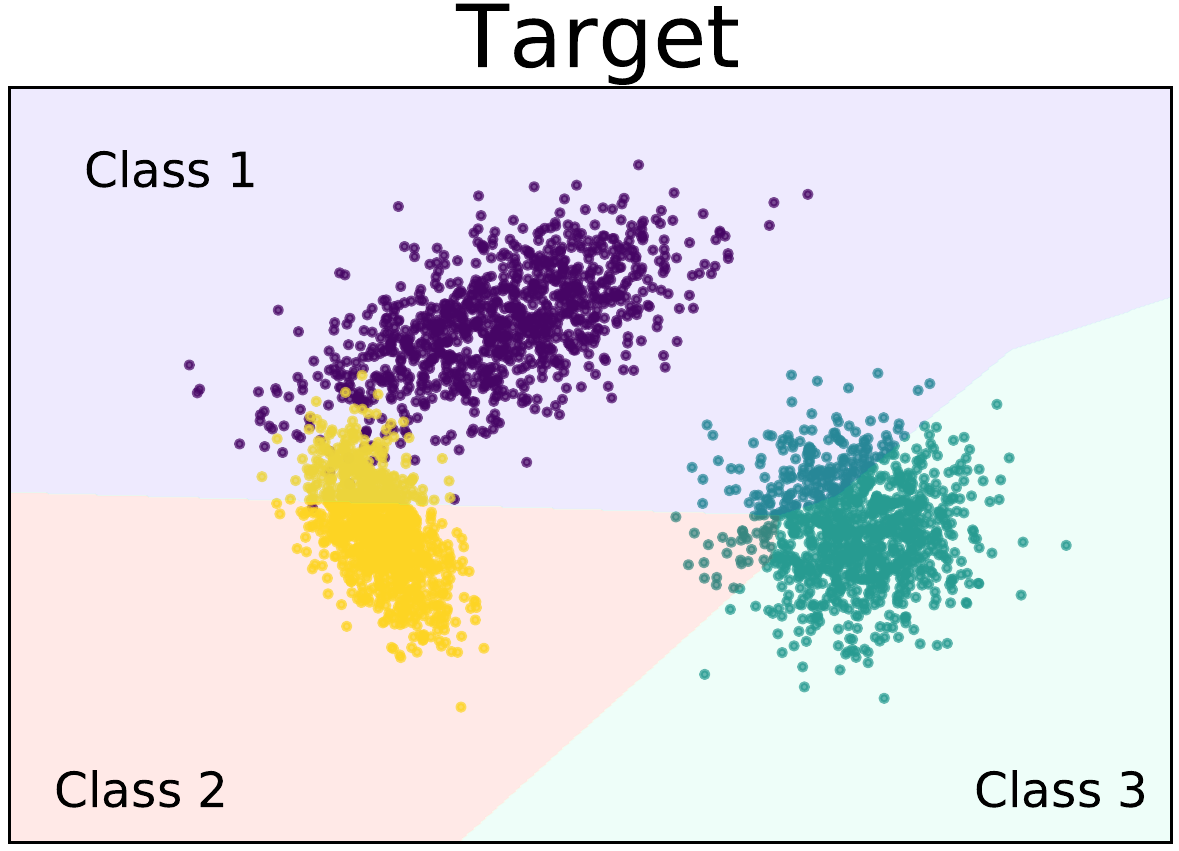}}
    \subfigure[\label{fig:conf_ent}]{\includegraphics[width=.186\linewidth]{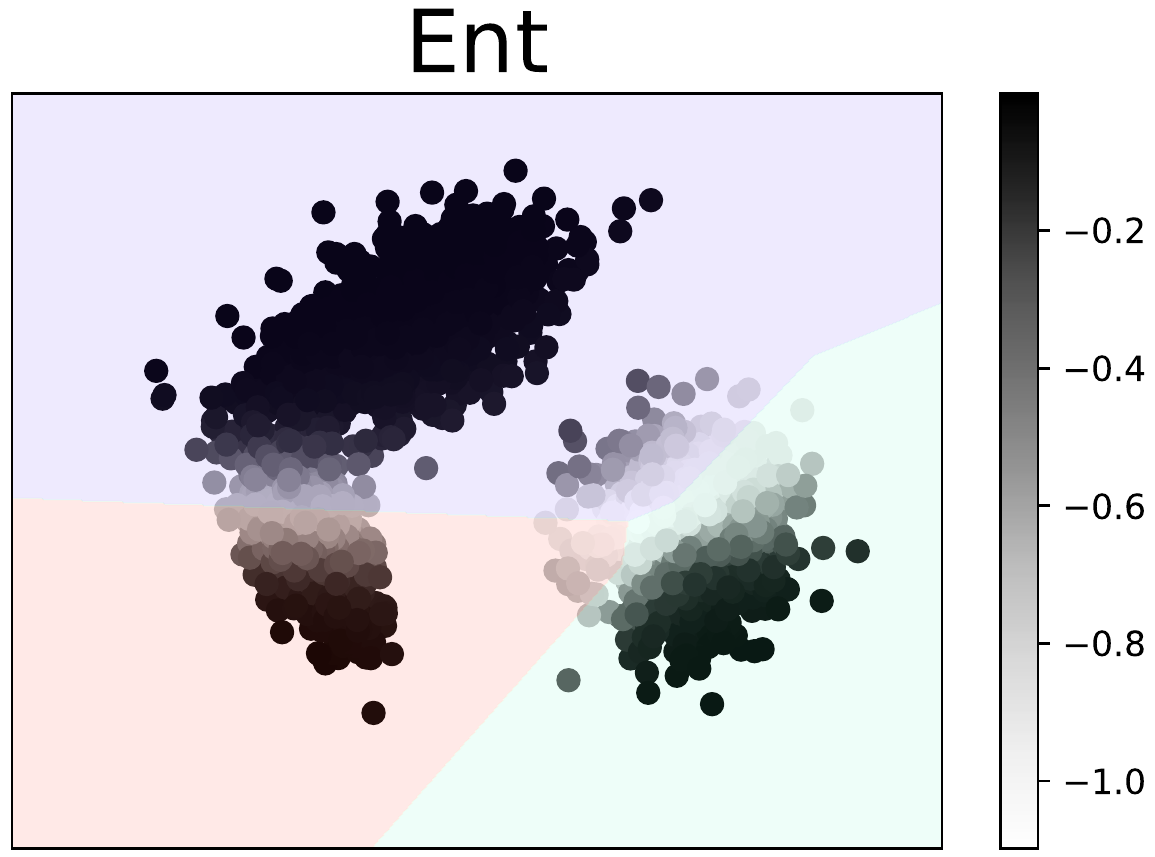}}
    \subfigure[\label{fig:conf_mppl_model}]{\includegraphics[width=.18\linewidth]{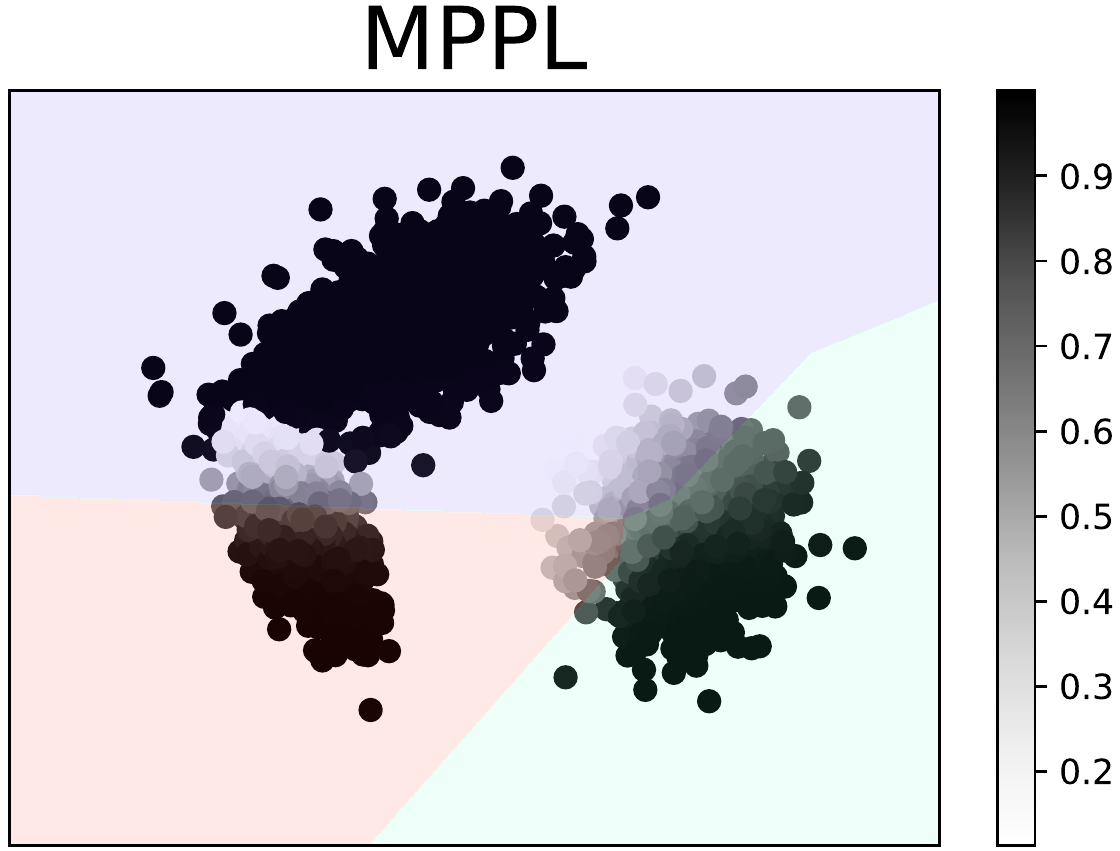}}
    \subfigure[\label{fig:conf_lpg_model}]{\includegraphics[width=.18\linewidth]{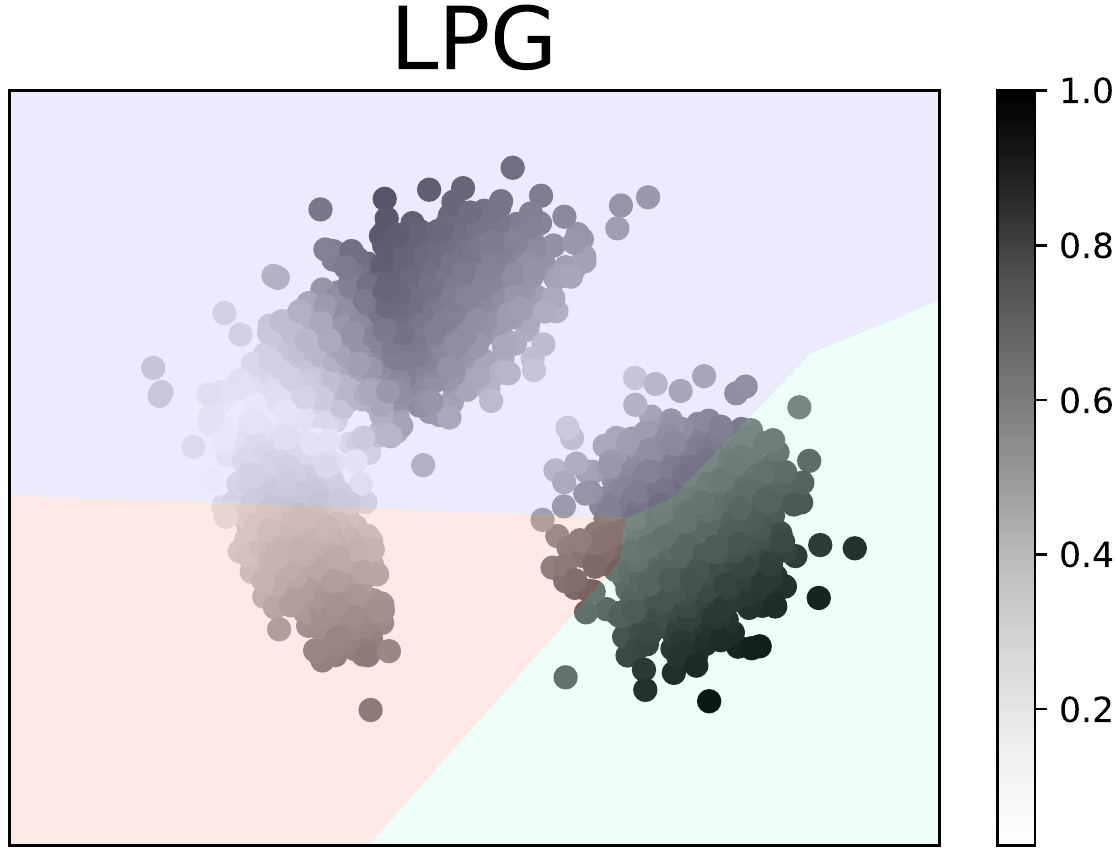}}
    \subfigure[\label{fig:conf_jmds_model}]{\includegraphics[width=.18\linewidth]{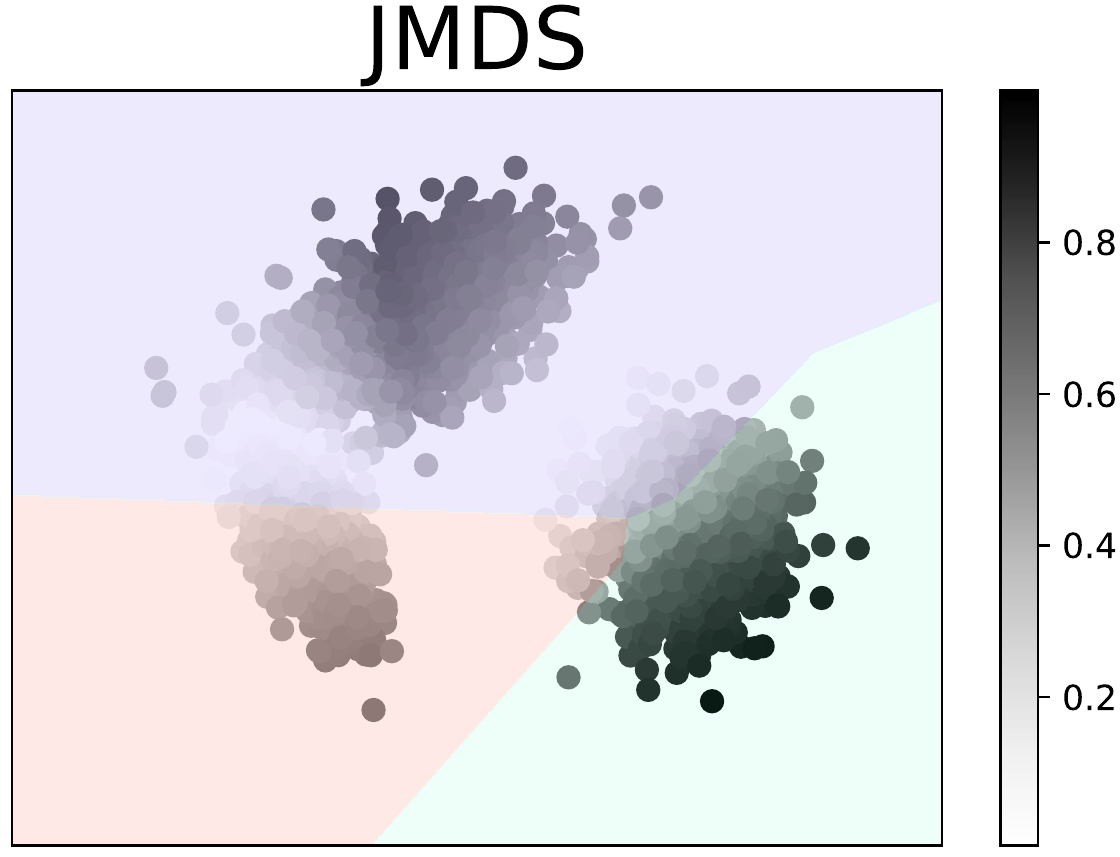}}
    \vspace{-0.325cm}
    \caption{
    A toy example to determine how the JMDS score works.
    (a) The source feature distribution. (f) The target feature distribution. 
    (b)-(e), (g)-(j) The higher the confidence score, the darker the point.
    (c)-(e) use the decision boundary of GMM and the others use the decision boundary of the pre-trained source model.
    }
    \label{fig:conf}
    \vspace{-0.125cm}
\end{figure*}

In this section, we analyze how the proposed components, the JMDS score and weight Mixup, work through additional experimental results.

\paragraph{JMDS score at the pre-trained source model:}

\begin{figure*}[t]
    \vspace{-0.2cm}
    \centering
    \subfigure[\label{fig:rccurve}]{\includegraphics[width=.23\linewidth]{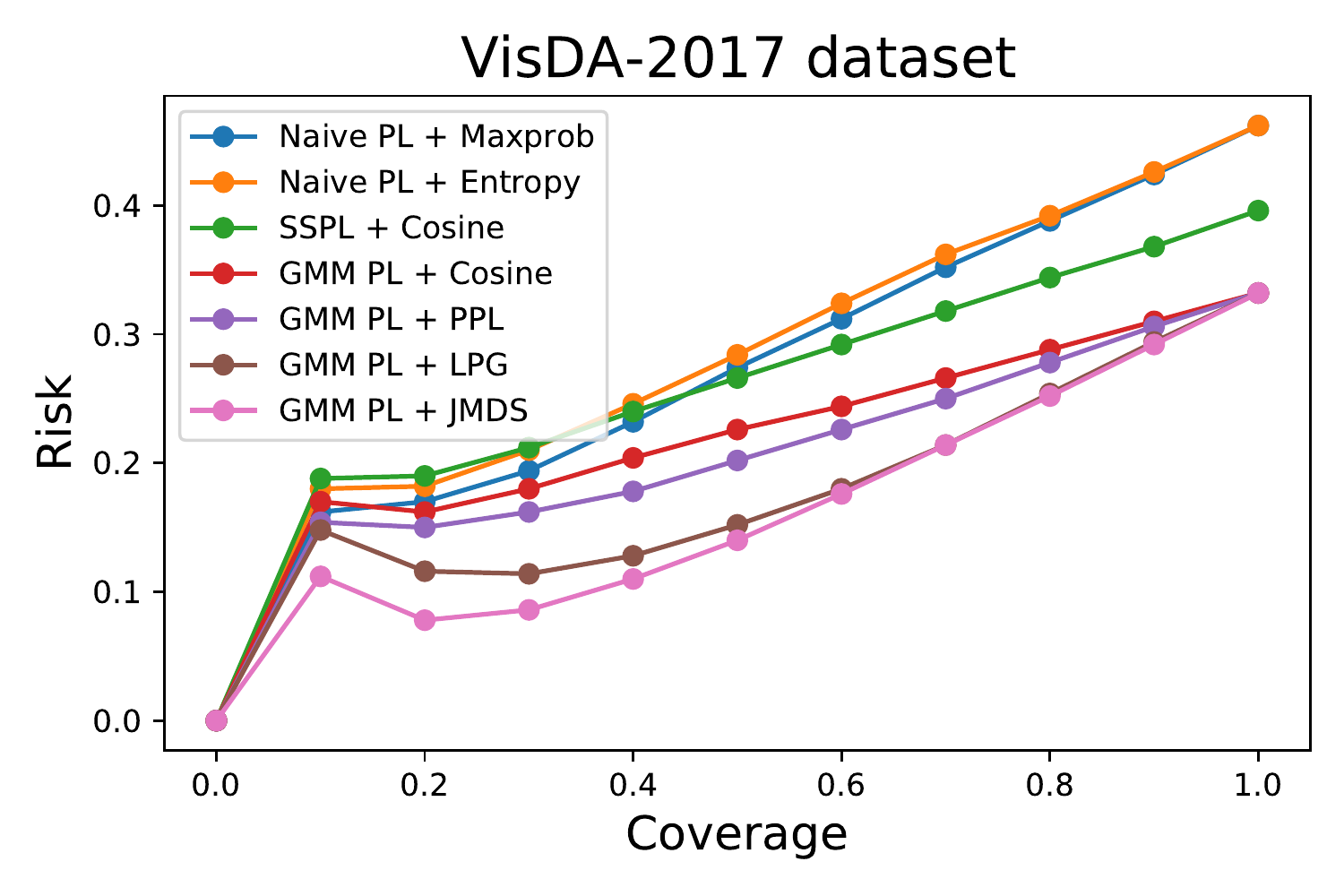}}
    \subfigure[\label{fig:comparison}]{\includegraphics[width=.23\linewidth]{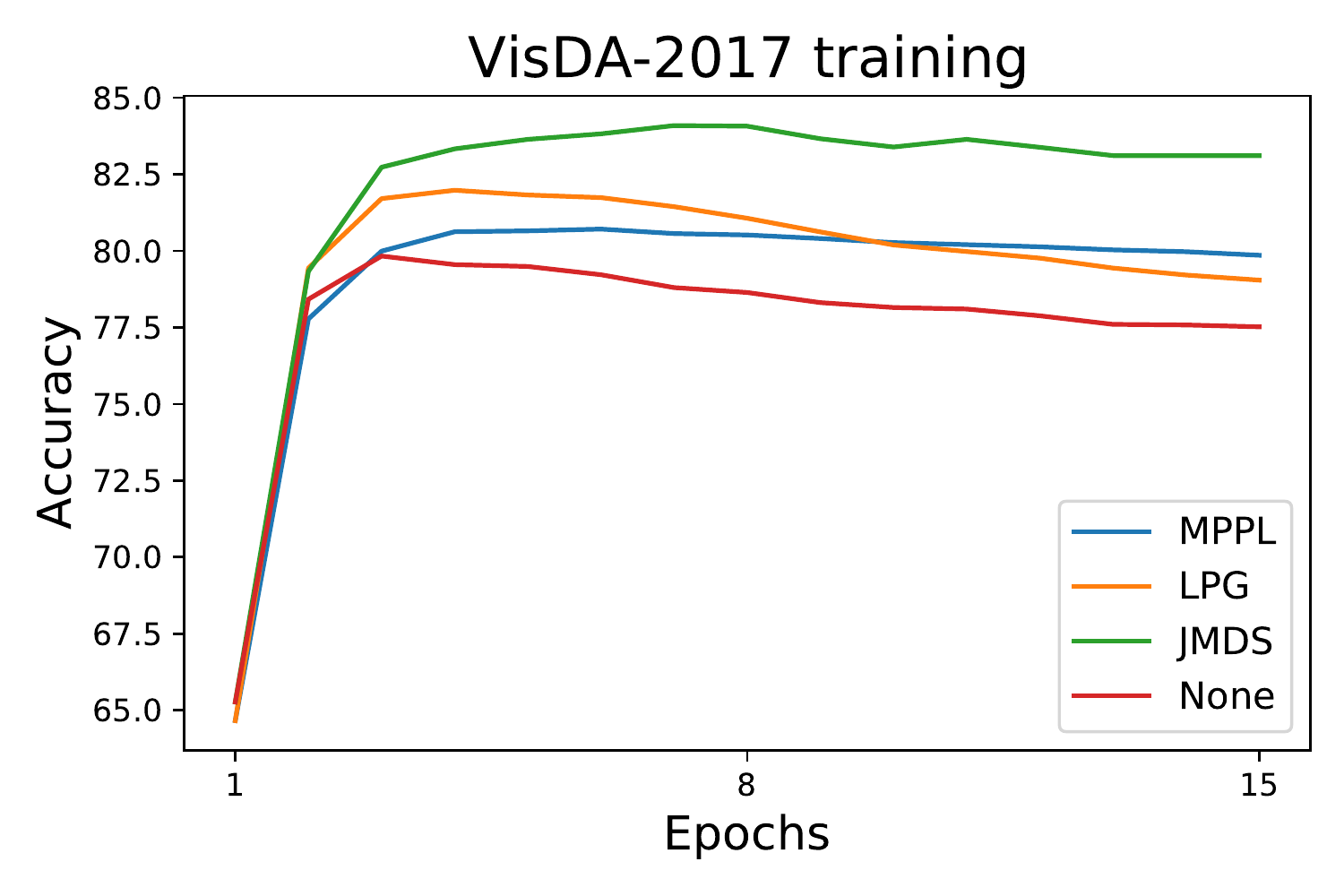}}
    \subfigure[\label{fig:quantile}]{\includegraphics[width=.23\linewidth]{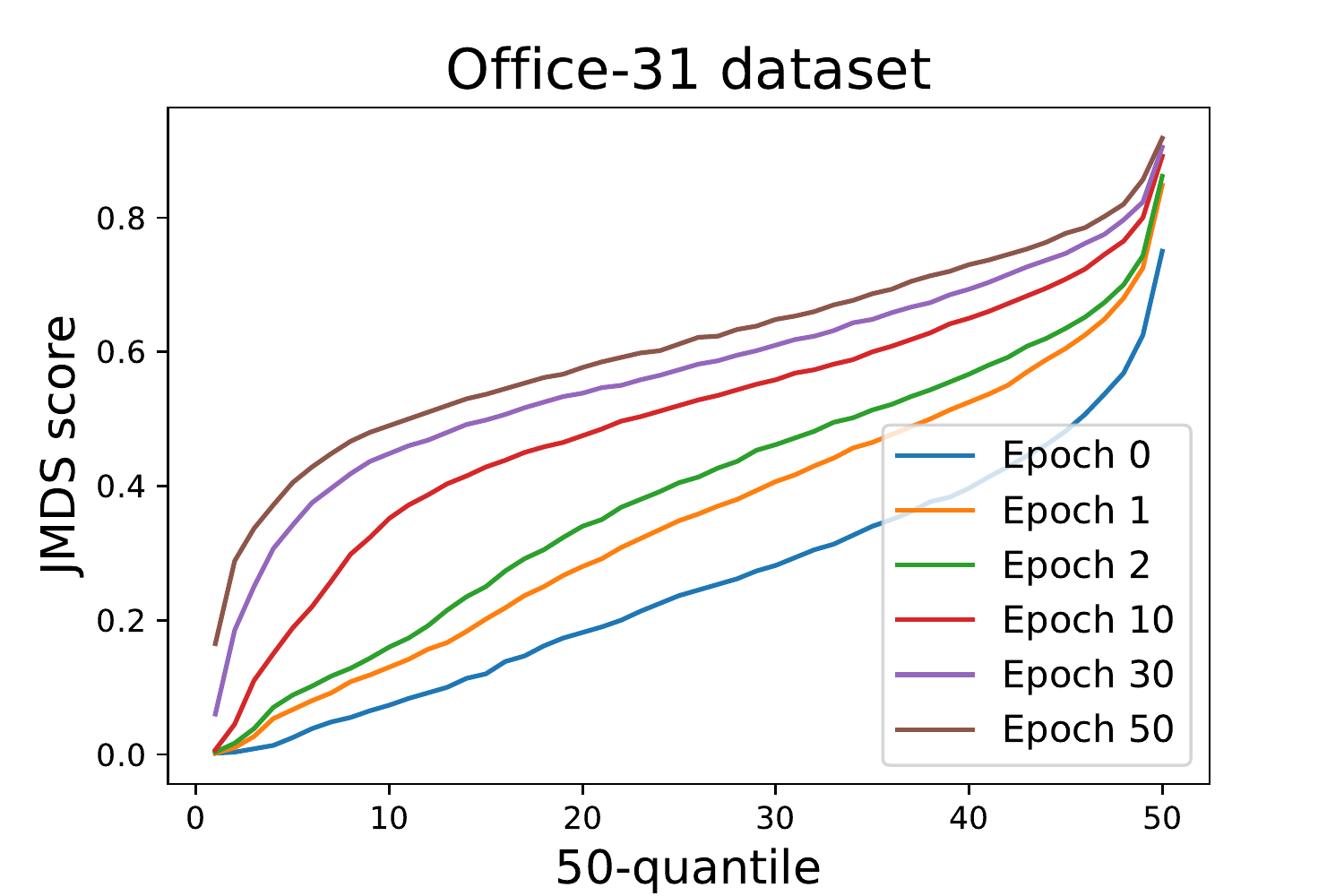}}
    \subfigure[\label{fig:weightmixupboost}]{\includegraphics[width=.23\linewidth]{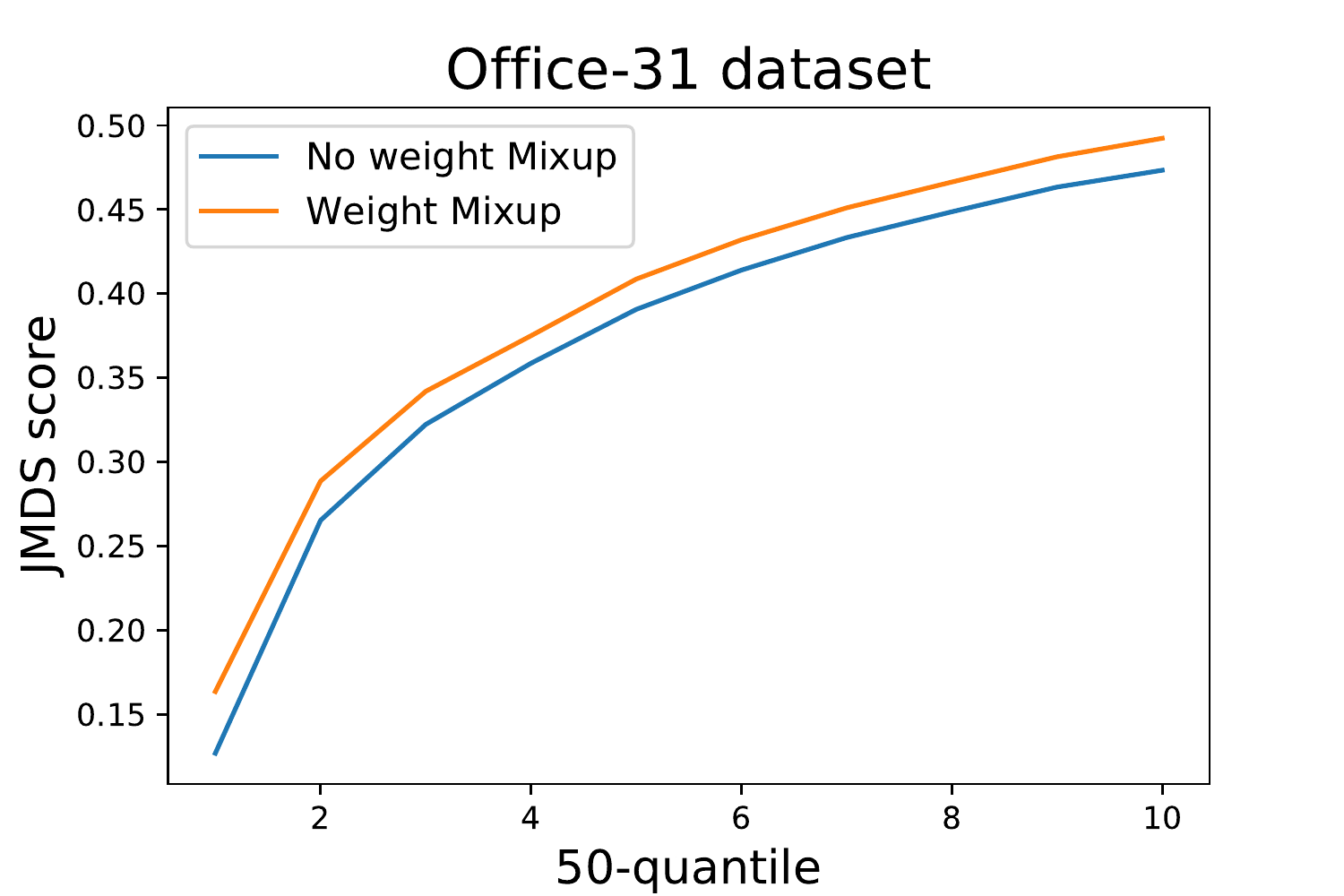}}
    \vspace{-0.3cm}
    \caption{(a) The RC curve of the pre-trained source model on the VisDA-2017 dataset. (b) The accuracy-epoch plot on the VisDA-2017 dataset. (c) The JMDS-quantile plot on various epochs. (d) The JMDS-quantile plot to demonstrate the effectiveness of weight Mixup.}
    \label{fig:discussion}
    \vspace{-0.35cm}
\end{figure*}

We considered a toy example to describe how the JMDS score works in Figure~\ref{fig:conf}.
Figure~\ref{fig:source_dist} and \ref{fig:target_dist} show the source and target data distributions, respectively.
We generated three-mode Gaussian source features and trained a two-layer fully-connected classifier.
Then, we generated another three-mode Gaussian target feature with a slightly different distribution than the source features and tested how various confidence scores are distributed.
Figure~\ref{fig:conf_maxprob},~\ref{fig:conf_ent}-\ref{fig:conf_jmds_model} follow the decision boundary of the model, and Figure~\ref{fig:conf_mppl_data}-\ref{fig:conf_jmds_data} follow the decision boundary of GMM.

The model-wise confidence score, Maxprob and Ent scores, have almost the same distribution as shown in Figure~\ref{fig:conf_maxprob} and \ref{fig:conf_ent}. 
Because these two scores used the decision boundary of the model, as shown in Figure~\ref{fig:target_dist}, there were many confidently misclassified samples.
In Figure~\ref{fig:conf_mppl_model}, when using MPPL scores, confusing samples, which are in an overlapped region have lower confidence scores than the aforementioned two scores.
However, MPPL scores still give overconfident scores for most of the samples, because DNNs using ReLU are overconfident for most of the data~\cite{overconfident}. It causes negative effects on training.

In Figure~\ref{fig:conf_lpg_data}, LPG scores work well for the decision boundary of GMM.
However, in Figure~\ref{fig:conf_lpg_model}, samples near the decision boundary of the model in class 3 have high confidence because they do not use model-wise knowledge.
This is undesirable to confidence scores because samples near the decision boundary of the model should be suppressed.
Based on the cluster assumption~\cite{clusterassumption}, the decision boundary of the model does not pass through the training data when they fit.
In other words, features near the decision boundary of the model are far from the training feature distribution, such as out-of-distribution samples.
Therefore, the model cannot provide precise predictions for these features; hence, they need to be suppressed.

By considering both model-wise and data-structure-wise knowledge, the JMDS score, which combines MPPL and LPG scores, can overcome the aforementioned problems of overconfidence and out-of-distribution problems.
As shown in Figure~\ref{fig:conf_jmds_data} and ~\ref{fig:conf_jmds_model}, the JMDS score shows a reliable distribution in terms of both model-wise and data-structure-wise knowledge.
Quantitatively, the JMDS score achieved the best AURC value in SFUDA as shown in Table~\ref{Table:JMDSeval}. 
Qualitatively, as shown in Figure~\ref{fig:rccurve}, the JMDS score has the lowest risk for any coverage at the pre-trained source model on the VisDA-2017 dataset.

\paragraph{JMDS score during learning:}

Figure~\ref{fig:conf} and Table~\ref{Table:JMDSeval} show that the JMDS score is a reasonable confidence score in the epoch 0 state, that is, without learning. It is necessary to check whether it is effective when it is applied as a sample-wise weight during learning. 
According to~\citet{reweightrobust}, one crucial advantage of sample weighting is robustness against training set bias such as label noise. Therefore, CoWA-JMDS shows robustness to incorrect pseudo-labels through sample weighting using the JMDS score so that it performs well in SFUDA.

As shown in Figure~\ref{fig:comparison}, sample weighting using confidence scores (LPG, MPPL, and JMDS) improved the performance compared to when sample weighting was not performed (None). Additionally, sample weighting using JMDS was better than when sample weighting was performed using MPPL or LPG alone. This proves that the JMDS score has a positive effect not only on the pre-trained source model but also on the actual SFUDA learning.

Figure~\ref{fig:conf_jmds_data} shows many samples had low JMDS values.
Low values can cause an underfitting problem during learning. However, as CoWA-JMDS training progressed, the JMDS scores also increased, shown in Figure~\ref{fig:quantile}.
It indicates that the participation of data in learning increases as learning continues.
This prevents the underfitting problem and allows CoWA-JMDS to learn properly using all samples without sample selection.
The result is related to a progressive learning scheme, such as curriculum learning~\cite{curriculumlearning} and self-paced learning~\cite{selfpacedlearning}.

\paragraph{Effect of weight Mixup:}
We proposed weight Mixup to exploit more knowledge of the target feature distribution with low-confidence samples during CoWA-JMDS.
Figure~\ref{fig:weightmixupboost} shows the effect of weight Mixup during CoWA-JMDS training.
Because low-confidence samples are indirectly more included in learning as a form of mixed mid-level confidence samples, weight Mixup boosts the JMDS score of low-confidence samples.
It encourages the model can utilize more knowledge of the target feature distribution than when weight Mixup is not used.

As shown in Table~\ref{Table:mixup}, na\"ive use of Mixup leads marginal improvement in terms of average accuracy. Mixup boosts 0.2\% and 0.3\% for SHOT and the pseudo-labels of GMM, respectively.
This is still due to the vulnerability of the confirmation bias problem because Mixup gives the same importance to the mixed samples without considering the incorrect pseudo-labels in SFUDA.
Therefore, we need to use weight Mixup in SFUDA.
Weight Mixup increases performance by 3.4\% compared to Mixup with the pseudo-labels of GMM which utilizes the same pseudo-labels.
Also, weight Mixup boosts 0.7\% when it is applied to CoWA-JMDS.
It demonstrates that weight Mixup is an effective technique that can be used in learning with sample weighting based on confidence scores.

\begin{table}[t]
    \vspace{-0.25cm}
    \small
    \caption{A weight mixup experiment on Office-31 dataset.}
    \label{Table:mixup}
    \centering
    \resizebox{0.9\linewidth}{!}
    {\begin{tabular}{lc}
        \toprule
        Method          & Avg. \\
        \midrule
        SHOT~\cite{shot}            & 88.6 \\
        SHOT + Mixup~\cite{mixup} & 88.8 \\
        GMM PL & 86.6 \\
        GMM PL + Mixup & 86.9\\
        GMM PL + JMDS & 89.6 \\
        GMM PL + JMDS + weight Mixup (CoWA-JMDS) & 90.3 \\
        
        \bottomrule                        
    \end{tabular}}
    \vspace{-0.375cm}
\end{table}

\section{Conclusion}
\label{7_Conclusion}
In this study, we propose the JMDS score, the novel confidence score for SFUDA to differentiate sample importance based on confidence for pseudo-labels.
The JMDS score jointly considers both of the source and target domain knowledge unlike existing scores.
We then propose CoWA-JMDS, the SFUDA method that uses the JMDS score as a sample-wise weight.
CoWA-JMDS also uses weight Mixup, which is the proposed variant of Mixup, to exploit more target domain knowledge.
CoWA-JMDS can be extended to more realistic scenarios, such as open-set and partial-set scenarios.
The experiments showed that the proposed JMDS score and CoWA-JMDS achieve state-of-the-art performance on UDA benchmarks in various SFUDA scenarios.

\section*{Acknowledgements}

This work was supported by the National Research Foundation of Korea (NRF) grant funded by the Korea government (MSIT) (No. 2022R1A3B1077720), Institute of Information \& communications Technology Planning \& Evaluation (IITP) grant funded by the Korea government(MSIT) [NO.2021-0-01343, Artificial Intelligence Graduate School Program (Seoul National University)], Institute of Information \& communications Technology Planning \& Evaluation (IITP) grant funded by the Korea government(MSIT) (2022-0-00959), AIRS Company in Hyundai Motor and Kia through HMC/KIA-SNU AI Consortium Fund, Samsung Electronics Co., Ltd. (Mobile Communications Business), and the BK21 FOUR program of the Education and Research Program for Future ICT Pioneers, Seoul National University in 2022.

\nocite{langley00}

\bibliography{example_paper}

\begin{thebibliography}{41}
\providecommand{\natexlab}[1]{#1}
\providecommand{\url}[1]{\texttt{#1}}
\expandafter\ifx\csname urlstyle\endcsname\relax
  \providecommand{\doi}[1]{doi: #1}\else
  \providecommand{\doi}{doi: \begingroup \urlstyle{rm}\Url}\fi

\bibitem[Arazo et~al.(2020)Arazo, Ortego, Albert, O’Connor, and
  McGuinness]{confirmationbias}
Arazo, E., Ortego, D., Albert, P., O’Connor, N.~E., and McGuinness, K.
\newblock Pseudo-labeling and confirmation bias in deep semi-supervised
  learning.
\newblock In \emph{2020 International Joint Conference on Neural Networks
  (IJCNN)}, pp.\  1--8. IEEE, 2020.

\bibitem[Bengio et~al.(2009)Bengio, Louradour, Collobert, and
  Weston]{curriculumlearning}
Bengio, Y., Louradour, J., Collobert, R., and Weston, J.
\newblock Curriculum learning.
\newblock In \emph{Proceedings of the 26th annual international conference on
  machine learning}, pp.\  41--48, 2009.

\bibitem[Cao et~al.(2018)Cao, Long, Wang, and Jordan]{pda}
Cao, Z., Long, M., Wang, J., and Jordan, M.~I.
\newblock Partial transfer learning with selective adversarial networks.
\newblock In \emph{Proceedings of the IEEE conference on computer vision and
  pattern recognition}, pp.\  2724--2732, 2018.

\bibitem[Ding et~al.(2020)Ding, Liu, Xiong, and Shi]{revisitingconfidencescore}
Ding, Y., Liu, J., Xiong, J., and Shi, Y.
\newblock Revisiting the evaluation of uncertainty estimation and its
  application to explore model complexity-uncertainty trade-off.
\newblock In \emph{Proceedings of the IEEE/CVF Conference on Computer Vision
  and Pattern Recognition Workshops}, pp.\  4--5, 2020.

\bibitem[Geifman \& El-Yaniv(2017)Geifman and El-Yaniv]{selective1}
Geifman, Y. and El-Yaniv, R.
\newblock Selective classification for deep neural networks.
\newblock \emph{arXiv preprint arXiv:1705.08500}, 2017.

\bibitem[Geifman et~al.(2018)Geifman, Uziel, and
  El-Yaniv]{defineconfidencescore}
Geifman, Y., Uziel, G., and El-Yaniv, R.
\newblock Bias-reduced uncertainty estimation for deep neural classifiers.
\newblock \emph{arXiv preprint arXiv:1805.08206}, 2018.

\bibitem[Grandvalet et~al.(2005)Grandvalet, Bengio, et~al.]{clusterassumption}
Grandvalet, Y., Bengio, Y., et~al.
\newblock Semi-supervised learning by entropy minimization.
\newblock \emph{CAP}, 367:\penalty0 281--296, 2005.

\bibitem[Gu et~al.(2020)Gu, Sun, and Xu]{RSDA-MSTN}
Gu, X., Sun, J., and Xu, Z.
\newblock Spherical space domain adaptation with robust pseudo-label loss.
\newblock In \emph{Proceedings of the IEEE/CVF Conference on Computer Vision
  and Pattern Recognition}, pp.\  9101--9110, 2020.

\bibitem[He et~al.(2016)He, Zhang, Ren, and Sun]{resnet}
He, K., Zhang, X., Ren, S., and Sun, J.
\newblock Deep residual learning for image recognition.
\newblock In \emph{Proceedings of the IEEE conference on computer vision and
  pattern recognition}, pp.\  770--778, 2016.

\bibitem[Hein et~al.(2019)Hein, Andriushchenko, and Bitterwolf]{overconfident}
Hein, M., Andriushchenko, M., and Bitterwolf, J.
\newblock Why relu networks yield high-confidence predictions far away from the
  training data and how to mitigate the problem.
\newblock In \emph{Proceedings of the IEEE/CVF Conference on Computer Vision
  and Pattern Recognition}, pp.\  41--50, 2019.

\bibitem[Hou \& Zheng(2021)Hou and Zheng]{sfit}
Hou, Y. and Zheng, L.
\newblock Visualizing adapted knowledge in domain transfer.
\newblock In \emph{Proceedings of the IEEE/CVF Conference on Computer Vision
  and Pattern Recognition}, pp.\  13824--13833, 2021.

\bibitem[Ioffe \& Szegedy(2015)Ioffe and Szegedy]{batchnorm}
Ioffe, S. and Szegedy, C.
\newblock Batch normalization: Accelerating deep network training by reducing
  internal covariate shift.
\newblock In \emph{International conference on machine learning}, pp.\
  448--456. PMLR, 2015.

\bibitem[Kang et~al.(2019)Kang, Jiang, Yang, and Hauptmann]{CAN}
Kang, G., Jiang, L., Yang, Y., and Hauptmann, A.~G.
\newblock Contrastive adaptation network for unsupervised domain adaptation.
\newblock In \emph{Proceedings of the IEEE Conference on Computer Vision and
  Pattern Recognition}, pp.\  4893--4902, 2019.

\bibitem[Kumar et~al.(2010)Kumar, Packer, and Koller]{selfpacedlearning}
Kumar, M.~P., Packer, B., and Koller, D.
\newblock Self-paced learning for latent variable models.
\newblock In \emph{NIPS}, volume~1, pp.\ ~2, 2010.

\bibitem[Lakshminarayanan et~al.(2016)Lakshminarayanan, Pritzel, and
  Blundell]{selective2}
Lakshminarayanan, B., Pritzel, A., and Blundell, C.
\newblock Simple and scalable predictive uncertainty estimation using deep
  ensembles.
\newblock \emph{arXiv preprint arXiv:1612.01474}, 2016.

\bibitem[Langley(2000)]{langley00}
Langley, P.
\newblock Crafting papers on machine learning.
\newblock In Langley, P. (ed.), \emph{Proceedings of the 17th International
  Conference on Machine Learning (ICML 2000)}, pp.\  1207--1216, Stanford, CA,
  2000. Morgan Kaufmann.

\bibitem[LeCun et~al.(2015)LeCun, Bengio, and Hinton]{lecun2015deep}
LeCun, Y., Bengio, Y., and Hinton, G.
\newblock Deep learning.
\newblock \emph{nature}, 521\penalty0 (7553):\penalty0 436--444, 2015.

\bibitem[Lee et~al.(2013)]{pseudolabel}
Lee, D.-H. et~al.
\newblock Pseudo-label: The simple and efficient semi-supervised learning
  method for deep neural networks.
\newblock In \emph{Workshop on challenges in representation learning, ICML},
  2013.

\bibitem[Lee et~al.(2018)Lee, Lee, Lee, and Shin]{mahalanobis}
Lee, K., Lee, K., Lee, H., and Shin, J.
\newblock A simple unified framework for detecting out-of-distribution samples
  and adversarial attacks.
\newblock \emph{Advances in neural information processing systems}, 31, 2018.

\bibitem[Li et~al.(2020)Li, Jiao, Cao, Wong, and Wu]{3cgan}
Li, R., Jiao, Q., Cao, W., Wong, H.-S., and Wu, S.
\newblock Model adaptation: Unsupervised domain adaptation without source data.
\newblock In \emph{Proceedings of the IEEE/CVF Conference on Computer Vision
  and Pattern Recognition}, pp.\  9641--9650, 2020.

\bibitem[Liang et~al.(2020{\natexlab{a}})Liang, Hu, and Feng]{shot}
Liang, J., Hu, D., and Feng, J.
\newblock Do we really need to access the source data? source hypothesis
  transfer for unsupervised domain adaptation.
\newblock In \emph{International Conference on Machine Learning}, pp.\
  6028--6039. PMLR, 2020{\natexlab{a}}.

\bibitem[Liang et~al.(2020{\natexlab{b}})Liang, Wang, Hu, He, and Feng]{baus}
Liang, J., Wang, Y., Hu, D., He, R., and Feng, J.
\newblock A balanced and uncertainty-aware approach for partial domain
  adaptation.
\newblock In \emph{Computer Vision--ECCV 2020: 16th European Conference,
  Glasgow, UK, August 23--28, 2020, Proceedings, Part XI 16}, pp.\  123--140.
  Springer, 2020{\natexlab{b}}.

\bibitem[Liu et~al.(2019)Liu, Cao, Long, Wang, and Yang]{sta}
Liu, H., Cao, Z., Long, M., Wang, J., and Yang, Q.
\newblock Separate to adapt: Open set domain adaptation via progressive
  separation.
\newblock In \emph{Proceedings of the IEEE/CVF Conference on Computer Vision
  and Pattern Recognition}, pp.\  2927--2936, 2019.

\bibitem[Luo et~al.(2020)Luo, Wang, Huang, and Baktashmotlagh]{pgl}
Luo, Y., Wang, Z., Huang, Z., and Baktashmotlagh, M.
\newblock Progressive graph learning for open-set domain adaptation.
\newblock In \emph{International Conference on Machine Learning}, pp.\
  6468--6478. PMLR, 2020.

\bibitem[Mandelbaum \& Weinshall(2017)Mandelbaum and Weinshall]{selective3}
Mandelbaum, A. and Weinshall, D.
\newblock Distance-based confidence score for neural network classifiers.
\newblock \emph{arXiv preprint arXiv:1709.09844}, 2017.

\bibitem[M{\"u}ller et~al.(2019)M{\"u}ller, Kornblith, and Hinton]{labelsmooth}
M{\"u}ller, R., Kornblith, S., and Hinton, G.
\newblock When does label smoothing help?
\newblock \emph{arXiv preprint arXiv:1906.02629}, 2019.

\bibitem[Na et~al.(2020)Na, Jung, Chang, and Hwang]{fixbi}
Na, J., Jung, H., Chang, H., and Hwang, W.
\newblock Fixbi: Bridging domain spaces for unsupervised domain adaptation.
\newblock \emph{arXiv preprint arXiv:2011.09230}, 2020.

\bibitem[Nair et~al.(2020)Nair, Precup, Arnold, and Arbel]{selective4}
Nair, T., Precup, D., Arnold, D.~L., and Arbel, T.
\newblock Exploring uncertainty measures in deep networks for multiple
  sclerosis lesion detection and segmentation.
\newblock \emph{Medical image analysis}, 59:\penalty0 101557, 2020.

\bibitem[Pan \& Yang(2009)Pan and Yang]{transferlearningsurvey}
Pan, S.~J. and Yang, Q.
\newblock A survey on transfer learning.
\newblock \emph{IEEE Transactions on knowledge and data engineering},
  22\penalty0 (10):\penalty0 1345--1359, 2009.

\bibitem[Panareda~Busto \& Gall(2017)Panareda~Busto and Gall]{oda}
Panareda~Busto, P. and Gall, J.
\newblock Open set domain adaptation.
\newblock In \emph{Proceedings of the IEEE International Conference on Computer
  Vision}, pp.\  754--763, 2017.

\bibitem[Peng et~al.(2017)Peng, Usman, Kaushik, Hoffman, Wang, and
  Saenko]{visda}
Peng, X., Usman, B., Kaushik, N., Hoffman, J., Wang, D., and Saenko, K.
\newblock Visda: The visual domain adaptation challenge.
\newblock \emph{arXiv preprint arXiv:1710.06924}, 2017.

\bibitem[Ren et~al.(2018)Ren, Zeng, Yang, and Urtasun]{reweightrobust}
Ren, M., Zeng, W., Yang, B., and Urtasun, R.
\newblock Learning to reweight examples for robust deep learning.
\newblock In \emph{International Conference on Machine Learning}, pp.\
  4334--4343. PMLR, 2018.

\bibitem[Saenko et~al.(2010)Saenko, Kulis, Fritz, and Darrell]{office31}
Saenko, K., Kulis, B., Fritz, M., and Darrell, T.
\newblock Adapting visual category models to new domains.
\newblock In \emph{European conference on computer vision}, pp.\  213--226.
  Springer, 2010.

\bibitem[Salimans \& Kingma(2016)Salimans and Kingma]{weightnorm}
Salimans, T. and Kingma, D.~P.
\newblock Weight normalization: A simple reparameterization to accelerate
  training of deep neural networks.
\newblock \emph{arXiv preprint arXiv:1602.07868}, 2016.

\bibitem[Tang et~al.(2020)Tang, Chen, and Jia]{srdc}
Tang, H., Chen, K., and Jia, K.
\newblock Unsupervised domain adaptation via structurally regularized deep
  clustering.
\newblock In \emph{Proceedings of the IEEE/CVF conference on computer vision
  and pattern recognition}, pp.\  8725--8735, 2020.

\bibitem[Van~der Maaten \& Hinton(2008)Van~der Maaten and Hinton]{tsne}
Van~der Maaten, L. and Hinton, G.
\newblock Visualizing data using t-sne.
\newblock \emph{Journal of machine learning research}, 9\penalty0 (11), 2008.

\bibitem[Venkateswara et~al.(2017)Venkateswara, Eusebio, Chakraborty, and
  Panchanathan]{officehome}
Venkateswara, H., Eusebio, J., Chakraborty, S., and Panchanathan, S.
\newblock Deep hashing network for unsupervised domain adaptation.
\newblock In \emph{Proceedings of the IEEE conference on computer vision and
  pattern recognition}, pp.\  5018--5027, 2017.

\bibitem[Xu et~al.(2019)Xu, Li, Yang, and Lin]{safn}
Xu, R., Li, G., Yang, J., and Lin, L.
\newblock Larger norm more transferable: An adaptive feature norm approach for
  unsupervised domain adaptation.
\newblock In \emph{Proceedings of the IEEE/CVF International Conference on
  Computer Vision}, pp.\  1426--1435, 2019.

\bibitem[Yang et~al.(2020)Yang, Wang, van~de Weijer, Herranz, and Jui]{bait}
Yang, S., Wang, Y., van~de Weijer, J., Herranz, L., and Jui, S.
\newblock Unsupervised domain adaptation without source data by casting a bait.
\newblock \emph{arXiv preprint arXiv:2010.12427}, 2020.

\bibitem[Yang et~al.(2021)Yang, van~de Weijer, Herranz, Jui, et~al.]{nrc}
Yang, S., van~de Weijer, J., Herranz, L., Jui, S., et~al.
\newblock Exploiting the intrinsic neighborhood structure for source-free
  domain adaptation.
\newblock \emph{Advances in Neural Information Processing Systems}, 34, 2021.

\bibitem[Zhang et~al.(2017)Zhang, Cisse, Dauphin, and Lopez-Paz]{mixup}
Zhang, H., Cisse, M., Dauphin, Y.~N., and Lopez-Paz, D.
\newblock mixup: Beyond empirical risk minimization.
\newblock \emph{arXiv preprint arXiv:1710.09412}, 2017.

\end{thebibliography}
\bibliographystyle{icml2022}

\newpage

\appendix
\onecolumn

\section{Previous confidence scores}
\label{Appendix:confidencescores}

We redefine the scores to be between [0,1].
\begin{equation*}
    \begin{gathered}
        \textmd{Maxprob}(x_i^{t}) = \max_c{p_M(x_i^{t})}_c, \\
        \textmd{Ent}(x_i^{t}) = 1 + \cfrac{\Sigma_{c=1}^{K} p_M(x_i^{t})_c\log{p_M(x_i^{t})_c}}{\log{K}}, \\
        \textmd{Cossim}(x_i^{t}) = \frac{1}{2}(1 + \frac{\langle x_i^{t},C_{\hat{y}_i^{t}} \rangle}{\Vert x_i^{t} \Vert \Vert C_{\hat{y}_i^{t}} \Vert}), \\
        \textmd{where} \; p(x)_c = p(\hat{y}=c|x), \\
        C_{\hat{y}_i^{t}} \; \textmd{is the center of the cluster corresponding to class } \hat{y}_i^{t}.
    \end{gathered}
\end{equation*}

\section{Gaussian Mixture Modeling (GMM)}
\label{Appendix:GMM}

Using GMM, we obtain the parameters $\mu_c, \Sigma_c$, and $\pi_c$ and log-likelihood 
\begin{equation}
\begin{gathered}
    \log{p(x_i^{t}|\mu_c,\Sigma_c)}= -\frac{1}{2} (d \log{2\pi} + \log{|\Sigma_c|} + \\ (f(x_i^{t})-\mu_c)^T \Sigma_c^{-1}(f(x_i^{t})-\mu_c)),
\end{gathered}
\label{eq:loglikelihood}
\end{equation}
where $\pi_c, \mu_c$ and $\Sigma_c$ are the mixing coefficient, mean vector and covariance matrix of the class $c \in \{1,2,\cdots,K\}$ respectively.
Now, we obtain the data-structure-wise probability ${p_{\textmd{data}}(x_i^{t})}_c$ for class $c$ based on the ratio of log-likelihood and the corresponding pseudo-label:
\begin{equation}
\begin{gathered}
    {p_{\textmd{data}}(x_i^{t})}_c=\frac{{\pi_c(x_i^{t})}{p(x_i^{t}|\mu_c,\Sigma_c)}}{\Sigma_{c'} \{{\pi_{c'}(x_i^{t})}{p(x_i^{t}|\mu_{c'},\Sigma_{c'})}\}} \\
    \textmd{where} \quad c,c' \in \{1,2,\cdots,K\}.\\
    \hat{y_i}=\argmax_{c}{{p_{\textmd{data}}(x_i^{t})}_c}
\end{gathered}
\label{eq:dataprob}
\end{equation}

\section{Algorithms}
\label{Appendix:algorithms}
Algorithm~\ref{alg:oda} shows the full process for known/unknown classification for an open-set scenario.
Algorithm~\ref{alg:pda} shows the full process for class estimation for a partial-set scenario.

\begin{algorithm}[!htb]
   \caption{Known/unknown classification}
   \label{alg:oda}
\begin{algorithmic}[1]
  \STATE {\bfseries Input:} Unlabeled target data $X_t$, the model $M=g \circ f$.
   \STATE Compute $p_M(X_t) = \textmd{softmax}\big(g(f(X_t))\big)$.    
   \STATE Compute the entropy of $p_M(X_t)$.
   \STATE Divide $X_t$ in the low-entropy cluster $C_{\textmd{low-entropy}}$ and the high-entropy cluster $C_{\textmd{high-entropy}}$ based on 2-class k-means.
   \STATE {\bfseries Output:} $C_{\textmd{low-entropy}}$.
\end{algorithmic}
\end{algorithm}

\begin{algorithm}[!htb]
\caption{Class estimation}
\label{alg:pda}
\begin{algorithmic}[1]
   \STATE {\bfseries Input:} Unlabeled target data $X_t$, the model $M=g \circ f$, and a threshold $\tau$.
   \STATE $C_{\textmd{partial}} = \{c_1, \dots , c_K\}$.
   \REPEAT
   \STATE Initialize $\pi, \mu, \Sigma$ based on $p_M(X_t)$ and $C_{\textmd{partial}}$.
   \STATE Perform one EM iteration for GMM.
   \STATE compute $p_\textmd{data}(X_t)$
   \FOR{$i=1$ {\bfseries to} $|C_{\textmd{partial}}|$}
   \IF{$\sum_{i=1}^{n_t}{p_\textmd{data}(x_i)_{c_j}} < \tau\cdot\frac{n_t}{|C_{\textmd{partial}}|}$}
   \STATE $C_{\textmd{partial}} \leftarrow C_{\textmd{partial}} \backslash c_j$
   \ENDIF
   \ENDFOR
   
   \UNTIL{$C_{\textmd{partial}}$ converges}
   \STATE {\bfseries Output:} $C_{\textmd{partial}}$
\end{algorithmic}
\end{algorithm}

\newpage

\section{Pseudo-code for CoWA-JMDS}
\label{Appendix:pseudocode}
Algorithm~\ref{alg:cowa} shows the full procedure of CoWA-JMDS. Full code is available at a supplementary file.

\begin{algorithm}[!htb]
\caption{CoWA-JMDS}
\label{alg:cowa}
\begin{algorithmic}[1]
   \STATE {\bfseries Input:} Unlabeled target data $X_t$, the model $M=g \circ f$.
   \STATE epoch $\leftarrow$ 0.
   \REPEAT
  \IF{Partial-set scenario}
  \STATE Perform class estimation.
  \ENDIF
   \STATE Perform GMM on $f(X_t)$ and compute $p_\textmd{data}(X_t)$.
  \IF{Open-set scenario}
  \STATE Perform known/unknown classification.
  \ENDIF
   \STATE Compute JMDS score using Equation~(\ref{eq:JMDS}).
   \FOR{$i \leftarrow 1$ {\bfseries to} $iterations\_per\_epoch$}
   \IF{No weight Mixup}
   \STATE Compute loss using Equation~(\ref{eq:CoWA-JMDS}).
   \ELSIF{Weight Mixup}
   \STATE Obtain mixed inputs $\tilde{x}^t$, pseudo-labels $\tilde{{y}}^t$, and JMDS scores $\tilde{w}^t$ using Equation~(\ref{eq:mixup}).
   \STATE Compute loss using Equation~(\ref{eq:CoWA-JMDS*}).
   \ENDIF
   \STATE Update the model $M$ using loss.
   \ENDFOR
   \STATE epoch $\leftarrow$ epoch$ + 1$.
   \UNTIL{$\textmd{epoch}<max\_epoch$}
\end{algorithmic}
\end{algorithm}

\section{Implementation details}
\label{Appendix:impdetail}

Office-31~\cite{office31} is a small-sized standard UDA benchmark with three domains from different sources, that is, collected from the Amazon website (A), Web camera (W), and DSLR (D).
The dataset contain 4,110 images of 31 object classes of office supplies.
Office-Home~\cite{officehome} is a medium-sized UDA benchmark that contains Artistic images (Ar), Clip Art (Cl), Product images (Pr), and Real-World images (Rw).
The dataset contain 15,500 images of 65 object classes.
VISDA-2017~\cite{visda} is a challenging large-sized UDA benchmark.
It contains a training dataset with 152,397 synthetic data and a test dataset with 55,388 real images with 12 categories. 

We use ResNet-50 or ResNet-101 \cite{resnet} pre-trained on the source data as our backbone network. Following Liang et al.\cite{shot}, we put a bottleneck layer with 256 units and a task-specific classifier layer. a weight normalization layer~\cite{weightnorm} is applied in last classifier layer and bottleneck layer consists batch normalization layer~\cite{batchnorm}. Pre-trained source model trained with label smoothing~\cite{labelsmooth}.
We use mini-batch SGD with momentum 0.9 and weight decay 1e-3 for all experiments. Learning rate of a bottleneck layer is 1e-2 and the remainders are 1e-3 for all three datasets: the Office-31, Office-Home, VisDA-2017 datasets. We do not use learning rate decay and the number of epochs are 50, 30, and 15, respectively. We set batch size 64 for all three benchmark datasets.
The hyperparameter of weight Mixup $\alpha$ is set to 0.2, 0.2, and 2.0, respectively.
The threshold $\tau$ for class estimation in a partial-set scenario is set to 0.3 for the Office-Home dataset.
Since we perform GMM in the high-dimensional feature space, Expectation-Maximization iteration is conducted once to resolve the instability.

\section{Hyperparameter sensitivity}
\label{Appendix:hypsen}
\begin{figure*}[!htb]
    \centering
    \subfigure[\label{fig:hypsen-31}]{\includegraphics[width=.3\linewidth]{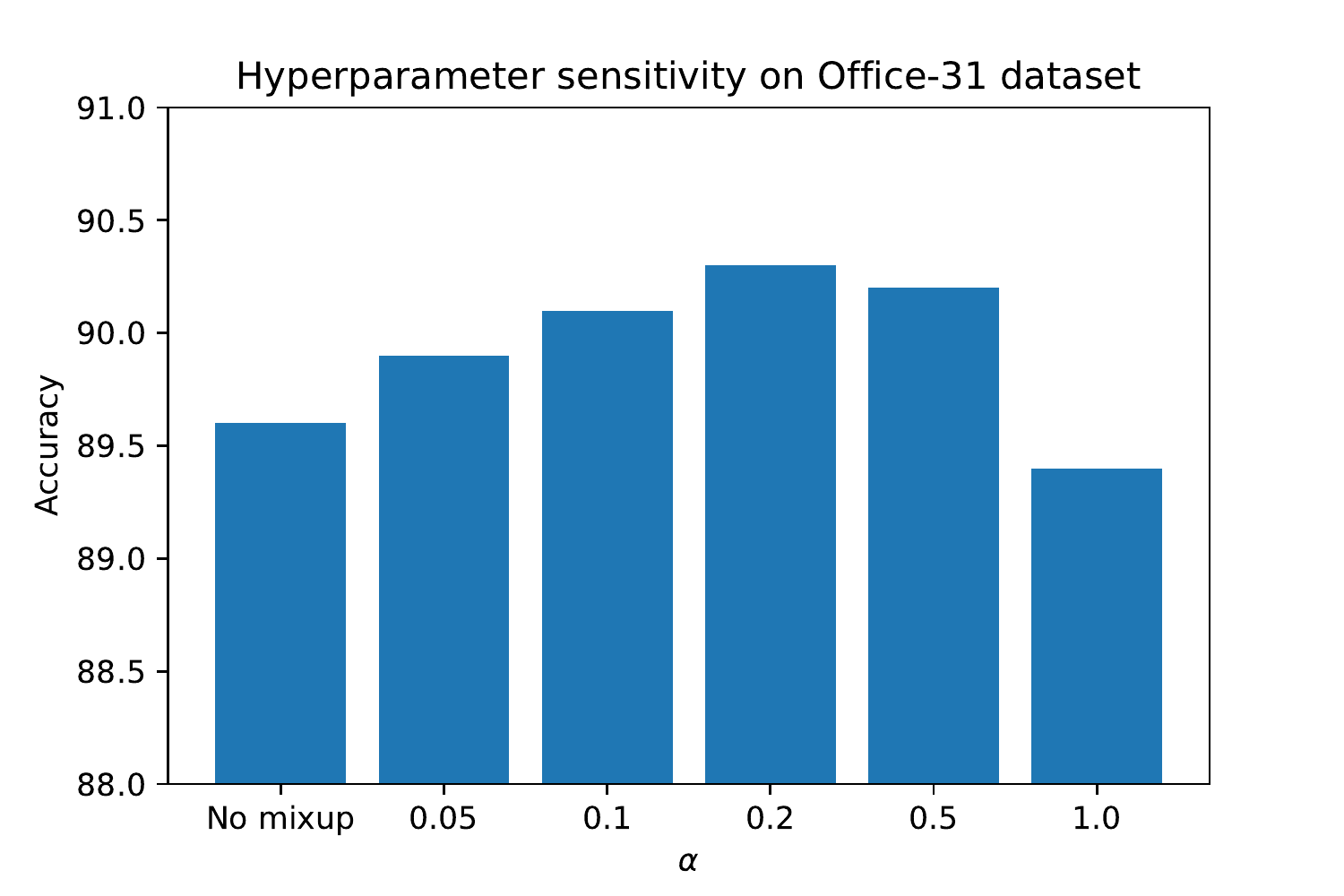}}
    \subfigure[\label{fig:hypsen_home}]{\includegraphics[width=.3\linewidth]{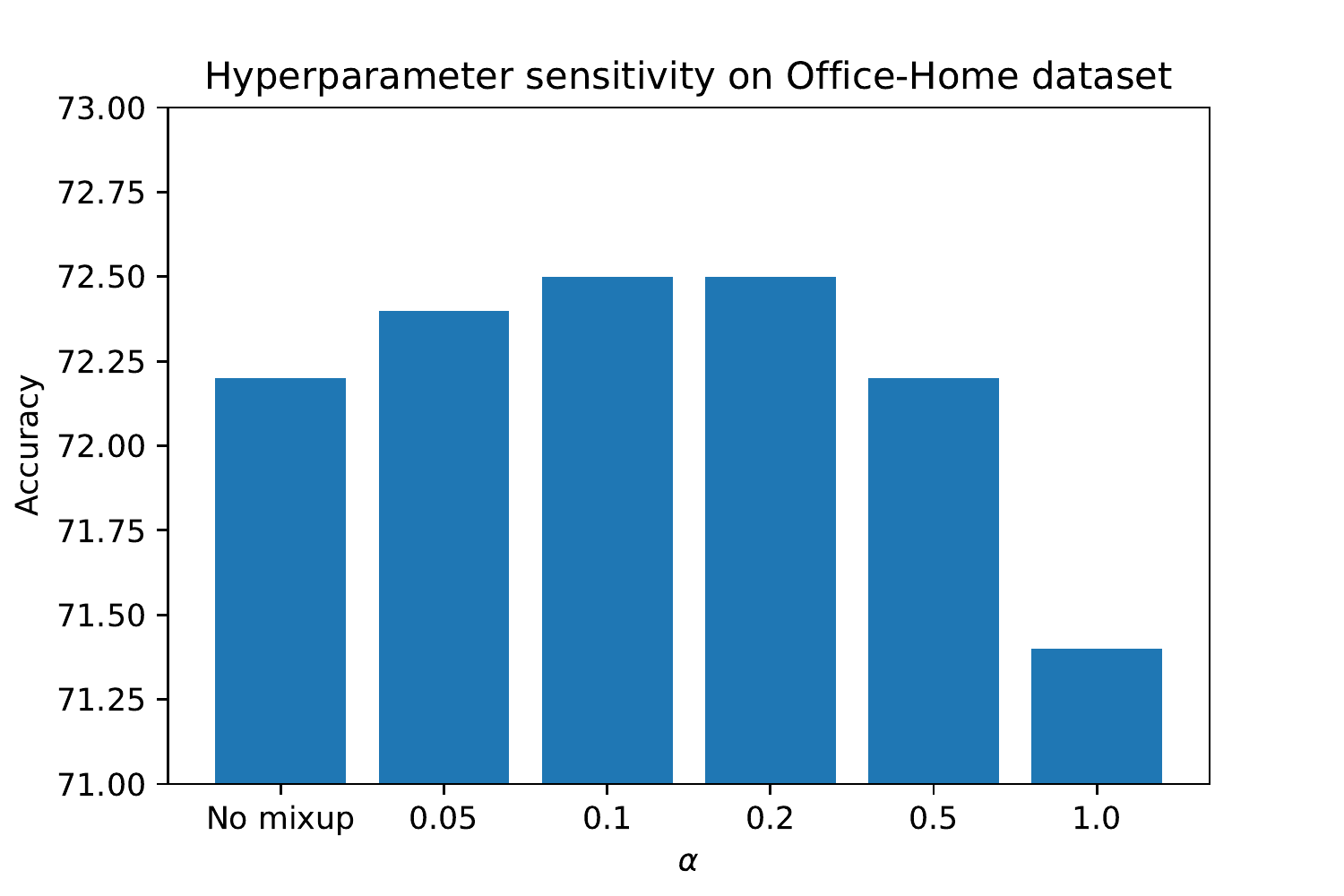}}
    \subfigure[\label{fig:hypsen_vis}]{\includegraphics[width=.3\linewidth]{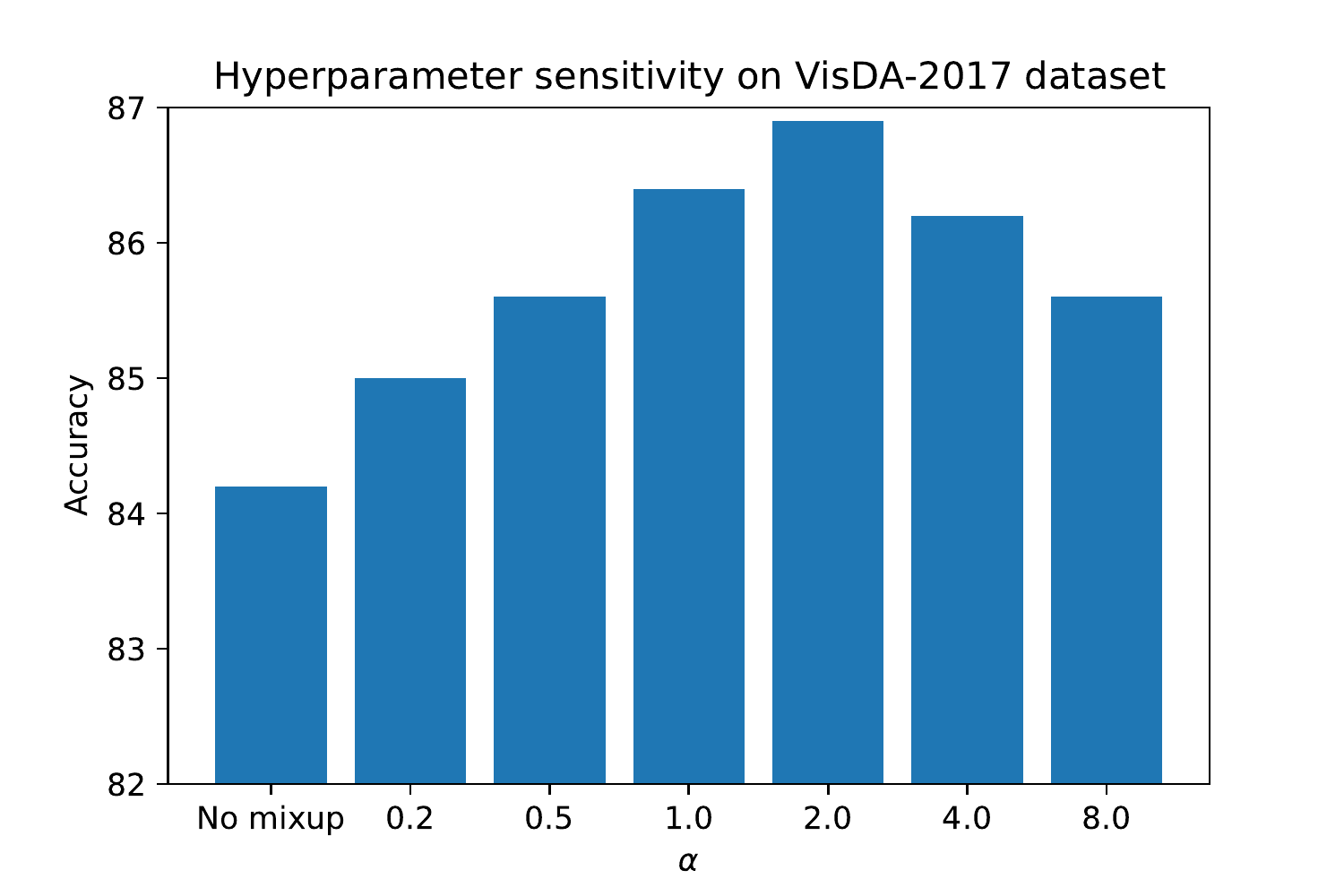}}
    \caption{Hyperparameter sensitivity result of $\alpha$ on three datasets.}
    \label{fig:hypsen}
\end{figure*}

Weight mixup has a hyperparameter $\alpha$ which decides the mixing coefficient $\gamma$.
Figure~\ref{fig:hypsen} shows the final accuracy of CoWA-JMDS for various values of $\alpha$.

\begin{figure*}[!htb]
    \centering
    \includegraphics[width=0.4\linewidth]{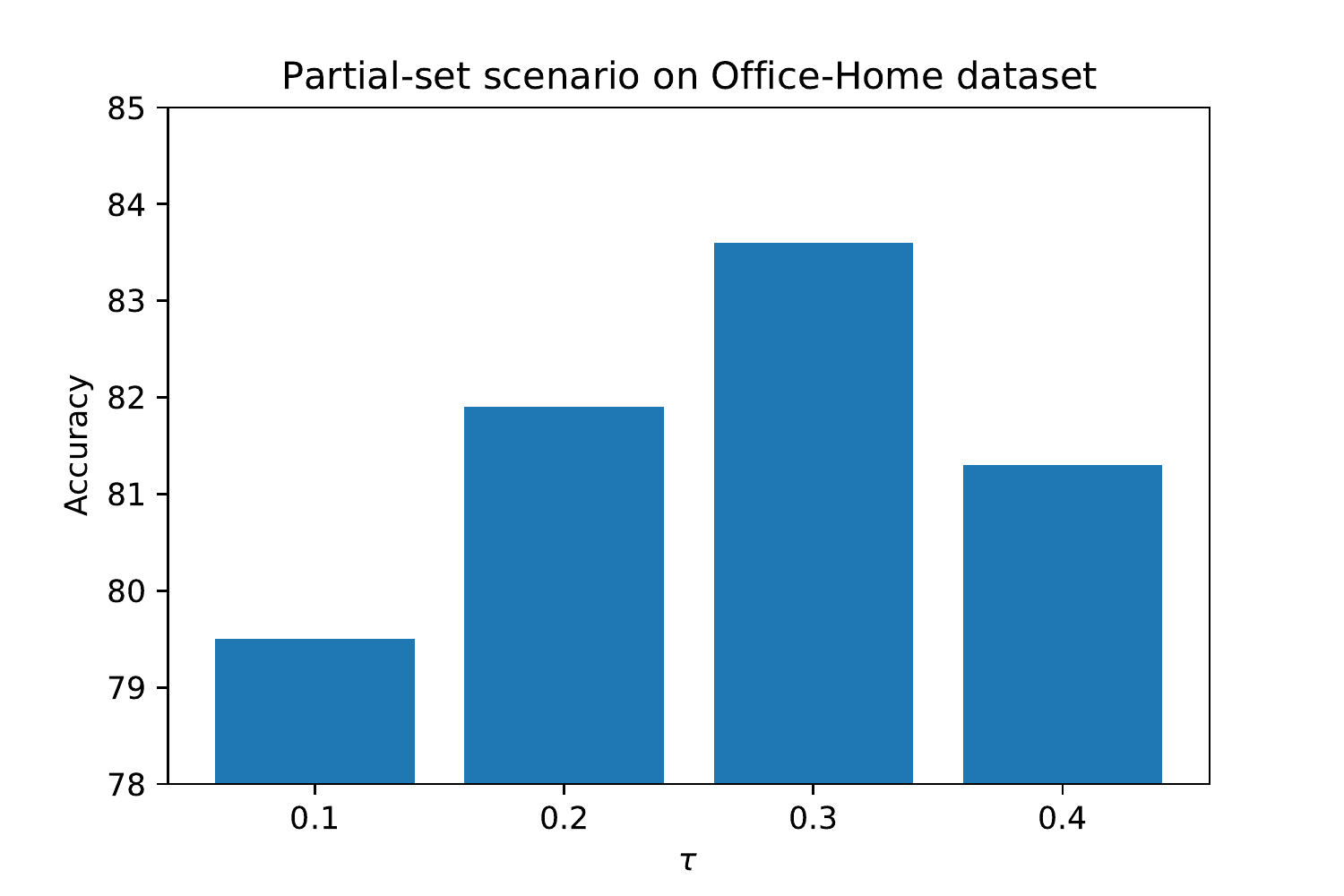}
    \vspace{-0.3cm}
    \caption{Hyperparameter sensitivity result of $\tau$ on the Office-Home dataset.
    }
    \label{fig:hypsen_tau}
    \vspace{-0.3cm}
\end{figure*}

In a partial-set scenario, there is a hyperparameter $\tau$ to estimate which classes are included in the target domain. Figure~\ref{fig:hypsen_tau} shows the accuracy of CoWA-JMDS for various values of $\tau$.

\end{document}